\documentclass[preprint,12pt]{elsarticle}

\usepackage{graphicx}%
\usepackage{algpseudocode}%
\usepackage{listings}%
\usepackage{amsmath,amsfonts}

\usepackage{float}
\usepackage{booktabs}
\usepackage{multirow}
\usepackage{makecell}
\usepackage{bbding}
\usepackage{adjustbox}
\usepackage[section]{placeins}
\usepackage{ulem}

\usepackage{lineno,hyperref}
\usepackage{threeparttable}
\usepackage{cleveref}
\usepackage{array}
\usepackage[caption=false,font=footnotesize,labelfont=rm,textfont=rm]{subfig}
\usepackage{textcomp}
\usepackage{stfloats}
\usepackage{url}
\usepackage{verbatim}
\usepackage{graphicx}
\usepackage{soul} 
\usepackage{longtable}
\usepackage{color, xcolor} 

\journal{Nuclear Physics B}

\begin{document}

\begin{frontmatter}

\title{Multispectral State-Space Feature Fusion: Bridging Shared and Cross-Parametric Interactions for Object Detection}

\author[a]{Jifeng~Shen\corref{mycorrespondingauthor}}
\cortext[mycorrespondingauthor]{Corresponding author}
\ead{shenjifeng@ujs.edu.cn}

\author[a]{Haibo~Zhan}
\author[b]{Shaohua~Dong}
\author[c]{Xin~Zuo}
\author[d]{Wankou~Yang}
\author[e]{Haibin~Ling}
\address[a]{School of Electrical and Information Engineering, Jiangsu University, Zhenjiang, 212013, China}
\address[b]{Department of Computer Science and Engineering, University of North Texas, Denton, TX 76207, USA}
\address[c]{School of Computer Science and Engineering, Jiangsu University of Science and Technology, Zhenjiang, 212003, China}
\address[d]{School of Automation, Southeast University, Nanjing, 210096, China}
\address[e]{Bodhi Intelligence Lab, Department of Artificial Intelligence, Westlake University, Hangzhou, Zhejiang 310030, China}

\begin{abstract}
Modern multispectral feature fusion for object detection faces two critical limitations: (1) Excessive preference for local complementary features over cross-modal shared semantics adversely affects generalization performance; and (2) The trade-off between the receptive field size and computational complexity present critical bottlenecks for scalable feature modeling. 
Addressing these issues, a novel Multispectral State-Space Feature Fusion framework, dubbed MS2Fusion, is proposed based on the state space model (SSM), achieving efficient and effective fusion through a dual-path parametric interaction mechanism. 
More specifically, the first cross-parameter interaction branch inherits the advantage of cross-attention in mining complementary information with cross-modal hidden state decoding in SSM. 
The second shared-parameter branch explores cross-modal alignment with joint embedding to obtain cross-modal similar semantic features and structures through parameter sharing in SSM.
Finally, these two paths are jointly optimized with SSM for fusing multispectral features in a unified framework, allowing our MS2Fusion to enjoy both functional complementarity and shared semantic space. 
Benefiting from the design of the dual-branch SSM, our approach simultaneously inherits the computational efficiency and the global receptive field, significantly improving the performance of multispectral object detection. 
Our experiments demonstrate that MS2Fusion consistently outperforms strong baselines: it improves mAP@0.5 by 3.3\% for object detection (FLIR), boosts mIoU by 2.8\% for RGB-T semantic segmentation (MFNet), and reduces MAE by 0.7\% for salient object detection (VT5000).
Notably, without requiring task-specific modifications, MS2Fusion achieves new state-of-the-art results across multiple multispectral perception tasks, demonstrating its remarkable generalization capability. The source code is available at https://github.com/61s61min/MS2Fusion.git.
\end{abstract}


\end{frontmatter}

\section{Introduction}
\label{sec1}
Multispectral object detection has recently drawn increasing interest owing to its robust performance by fusing information from multiple spectral bands, such as RGB and thermal bands. RGB images usually offer high resolution, rich color and texture features, but suffer from sharply performance deterioration in complex scenarios such as low light, adverse weather or occlusions. 
In contrast, thermal images can effectively overcome these environmental limitations but exhibit obvious deficiencies in color and texture details. 
Current single-modal generic object detection methods struggle to overcome these aforementioned challenges. However, multispectral feature fusion paves a way to provide a reliable object detection solution under such challenging conditions. 

\begin{figure*}[t]
	\centering
	\subfloat[ ]{\includegraphics[height=0.36\linewidth]{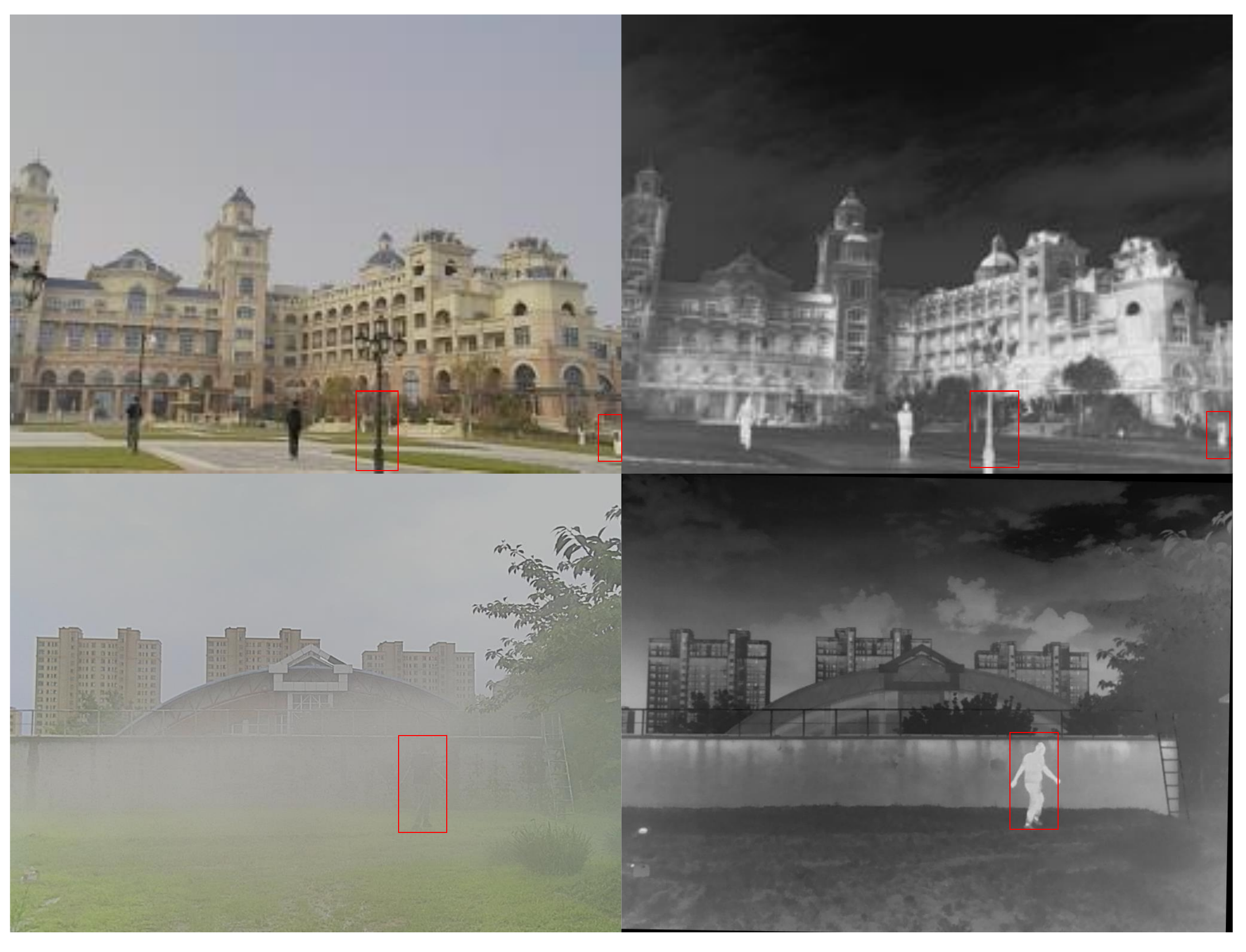}%
		\label{rgb_ir(a)}}
	\hfil
	\subfloat[ ]{\includegraphics[height=0.36\linewidth]{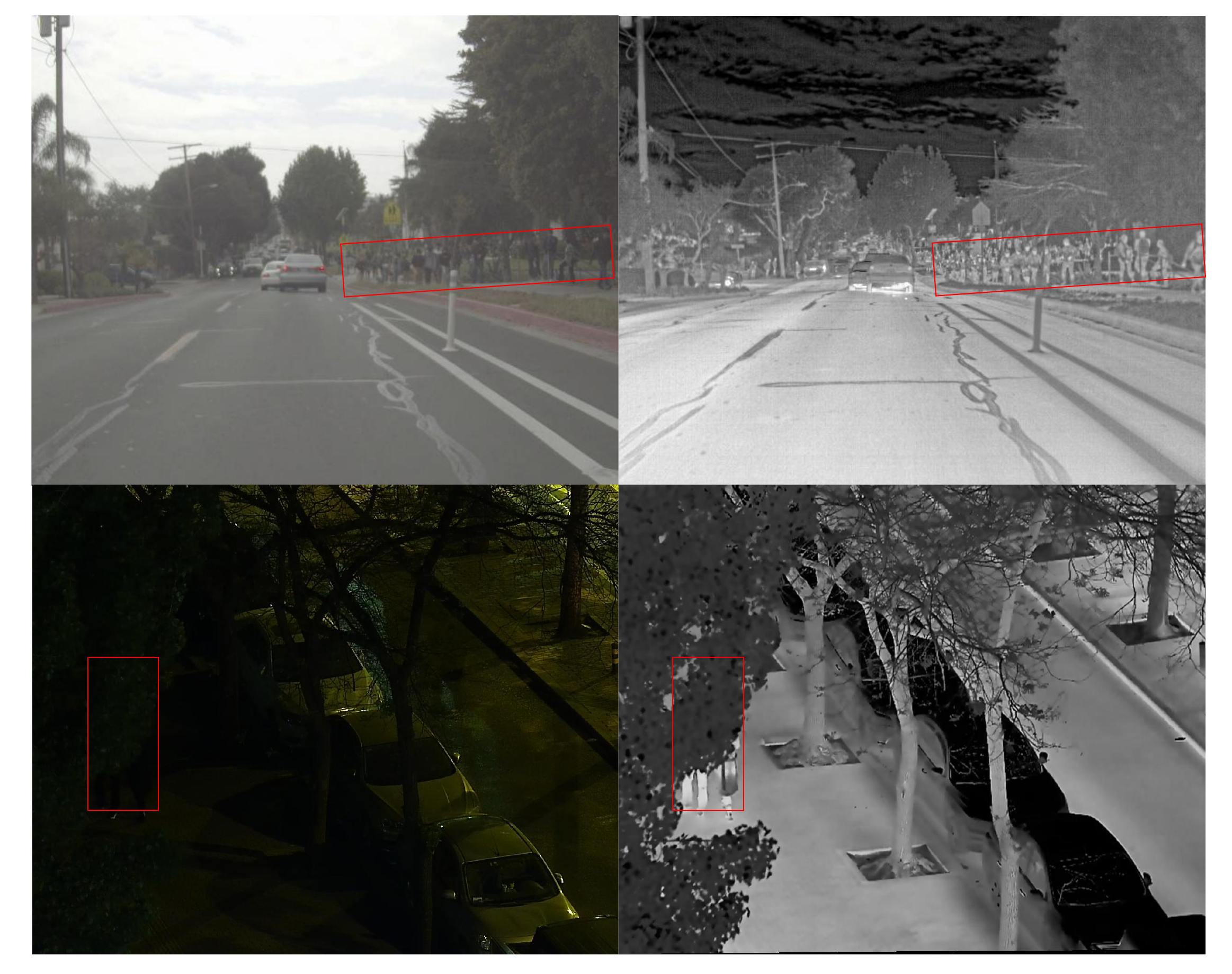}%
		\label{rgb_ir(b)}}
	\caption{The pros and cons of RGB (left) and thermal (right) images. (a) Both modalities provide complementary information, and their fusion enables more robust object detection; (b) Dual-modal shared features become crucial, since neither modality stands out distinctly. (e.g., modality-specific characteristics such as texture and thermal radiation are blurred, and cross-modal consistent features like object contours and structures are helpful for detection.)}
	\label{rgb_ir}
\end{figure*}

As shown in \Cref{rgb_ir(a)}, existing studies generally suggest that complementary features play a critical role when one modality is insufficient. For example, RGB images provide discriminative color and texture cues when thermal objects lack distinct contours, while thermal images offer thermal signatures when RGB imaging suffers from low illumination or occlusion. 
However, as demonstrated in \Cref{rgb_ir(b)}, scenarios where both modalities exhibit weak discriminative features (e.g., blurred textures in RGB and low contrast in thermal) cannot be resolved by complementary information alone. 
In such cases, shared features, such as cross-modal consistent shapes and structural patterns, become essential, as they capture modality-invariant representations for reliable detection. Thus, we speculate that a robust multispectral object detection framework should dynamically leverage both complementary and shared features to handle diverse real-world challenges.

Previous studies~\citep{liu2016multispectral, qingyun2021cross, 2021Multi, zhou2022unified, shin2024complementary, PENG2024410, 10234530}
have predominantly focused on complementary feature learning across modalities, often overlooking the exploration of shared feature representations or inherent structural similarities between them.
Moreover, existing approaches typically employ a single fusion strategy to directly combine multi-modal inputs, neglecting the potential benefits of adaptive or hierarchical fusion mechanisms.
These methods do not fully explore or exploit cross-modal shared features, ignoring the potential effect of enhancing the performance of single-modal features. 
This fusion paradigm often suppresses weaker yet discriminative features during cross-modal integration, leading to significant information loss. 
For multispectral object detection, shared feature representation plays a pivotal role in multi-modal fusion. Not only does it mitigate cross-modal discrepancies, but it also augments single-modal features, thereby substantially improving feature expressiveness and detection robustness in challenging environments.

In addition, most of the mainstream methods leverage CNN \citep{liu2016multispectral} or Transformer \citep{shen2024icafusion, dosovitskiy2020image, qingyun2021cross} for feature fusion. Albeit effective, existing CNN-based methods often struggle with capturing broader contextual information across modalities due to their limited receptive fields. On the other hand, despite excellence at modeling global dependencies, the Transformer-based approaches may degrade with longer input sequences, leading to performance degradation and higher computational costs as model complexity. These factors restrict their practicality in resource-constrained environments, underscoring the need for more efficient fusion strategies in multispectral object detection. 

Despite recent advances in multispectral object detection, existing methods predominantly focus on learning complementary features across modalities while neglecting the exploration of shared feature representations and inherent structural similarities. Most approaches adopt a single, direct fusion strategy, which often suppresses discriminative yet weaker features and leads to significant information loss. Moreover, while CNN-based methods struggle with limited receptive fields for capturing cross-modal context, Transformer-based techniques suffer from computational inefficiency and performance degradation with long sequences.  In speech recognition \citep{abdulaziz2023vowels} and image compression \citep{dawood2022rlc}, the efficiency of feature representation directly determines system performance - a principle equally applicable to multispectral detection. These limitations highlight the need for a more adaptive and efficient fusion paradigm that not only bridges modal discrepancies but also enhances single-modal features through shared representations, ultimately improving robustness in challenging environments.

\begin{figure*}[t]
	\centering
	\includegraphics[width=0.8\linewidth]{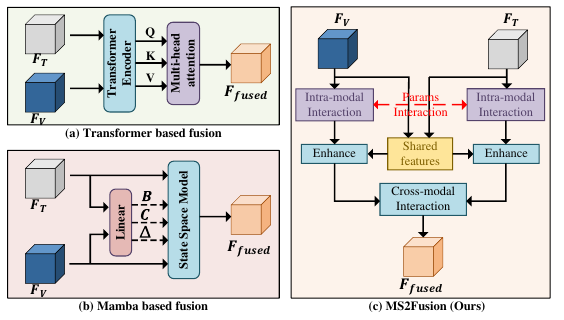}
	\caption{Comparing Transformer-based fusion (a), Mamba-based fusion (b) and our proposed MS2Fusion (c), with $F_T$ and $F_V$ as the input thermal and RGB image features, respectively. In Transformer-based method (a), $F_T$ and $F_V$ are fused through the multi-head attention mechanism, effectively integrating complementary information and enhancing performance across scenarios. The traditional Mamba approach (b) directly mixes dual-modal features to generate \textbf{B}, \textbf{C}, and \textbf{$\Delta$} parameters for SSM-based feature interaction, which may lead to modal misalignment and feature redundancy. In contrast, our method (c) first performs intra-modal feature interaction and then extracts cross-modal shared features, achieving better modal alignment and fusion, thereby providing more robust and unified feature representations.}
	\label{dif_compare}
\end{figure*}

\begin{figure*}[t]
	\centering
	\subfloat[CNN-based fusion]{\includegraphics[width=0.33\linewidth]{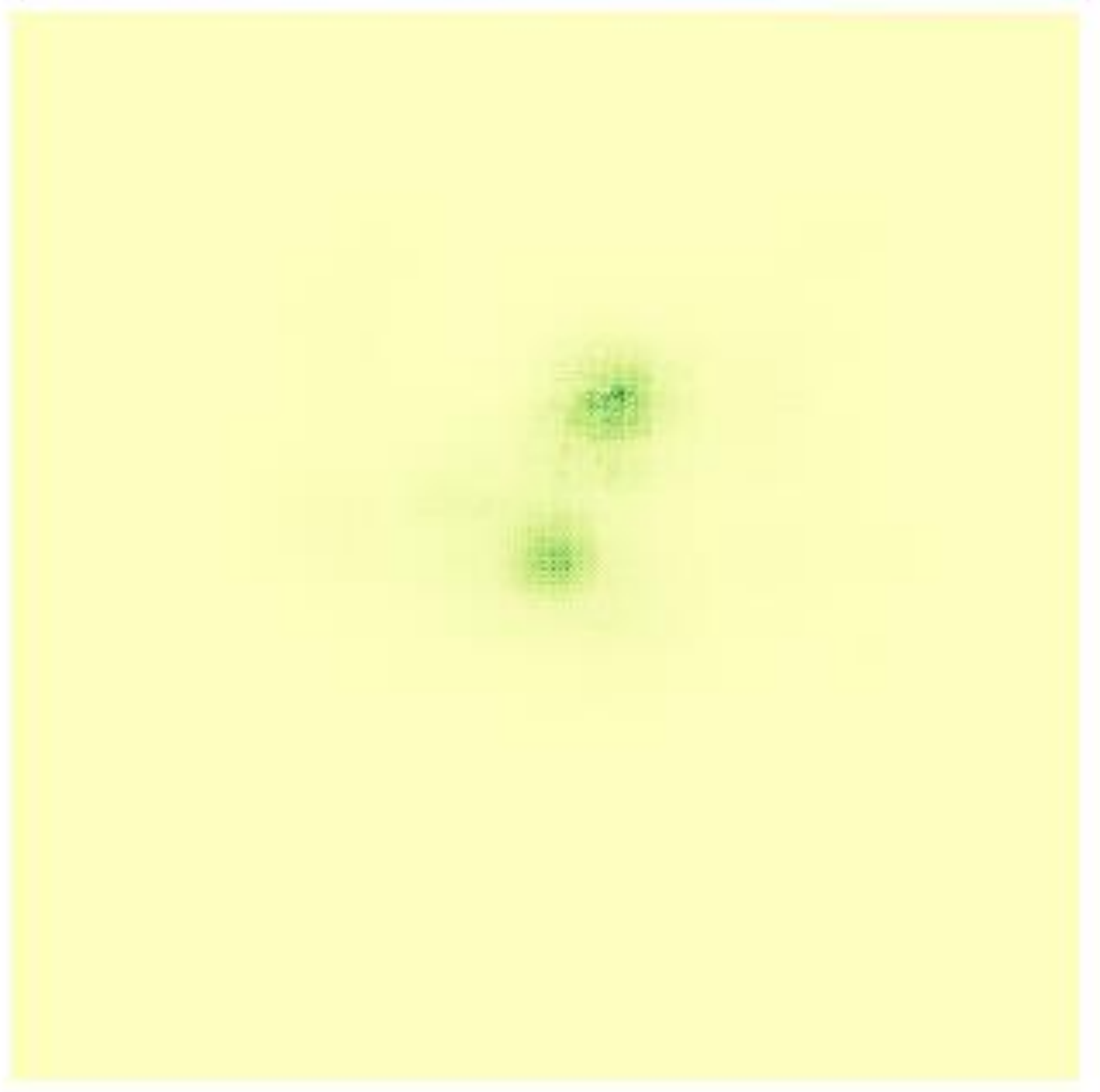}%
		\label{cnn_erf}}
	\hfil
	\subfloat[Transfomer-based fusion]{\includegraphics[width=0.33\linewidth]{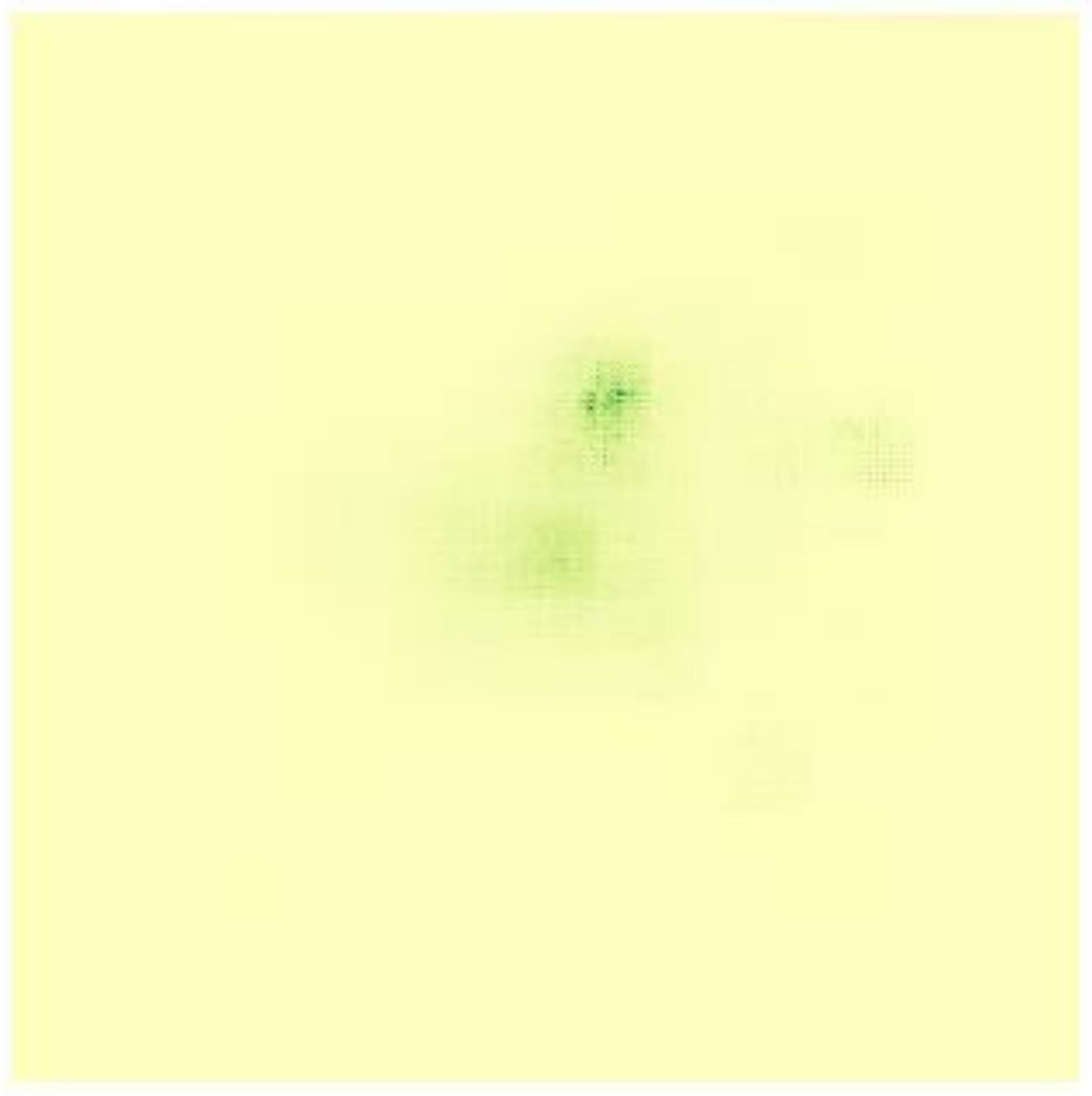}%
		\label{transformer_erf}}
        \hfil
        \subfloat[MS2Fusion]{\includegraphics[width=0.33\linewidth]{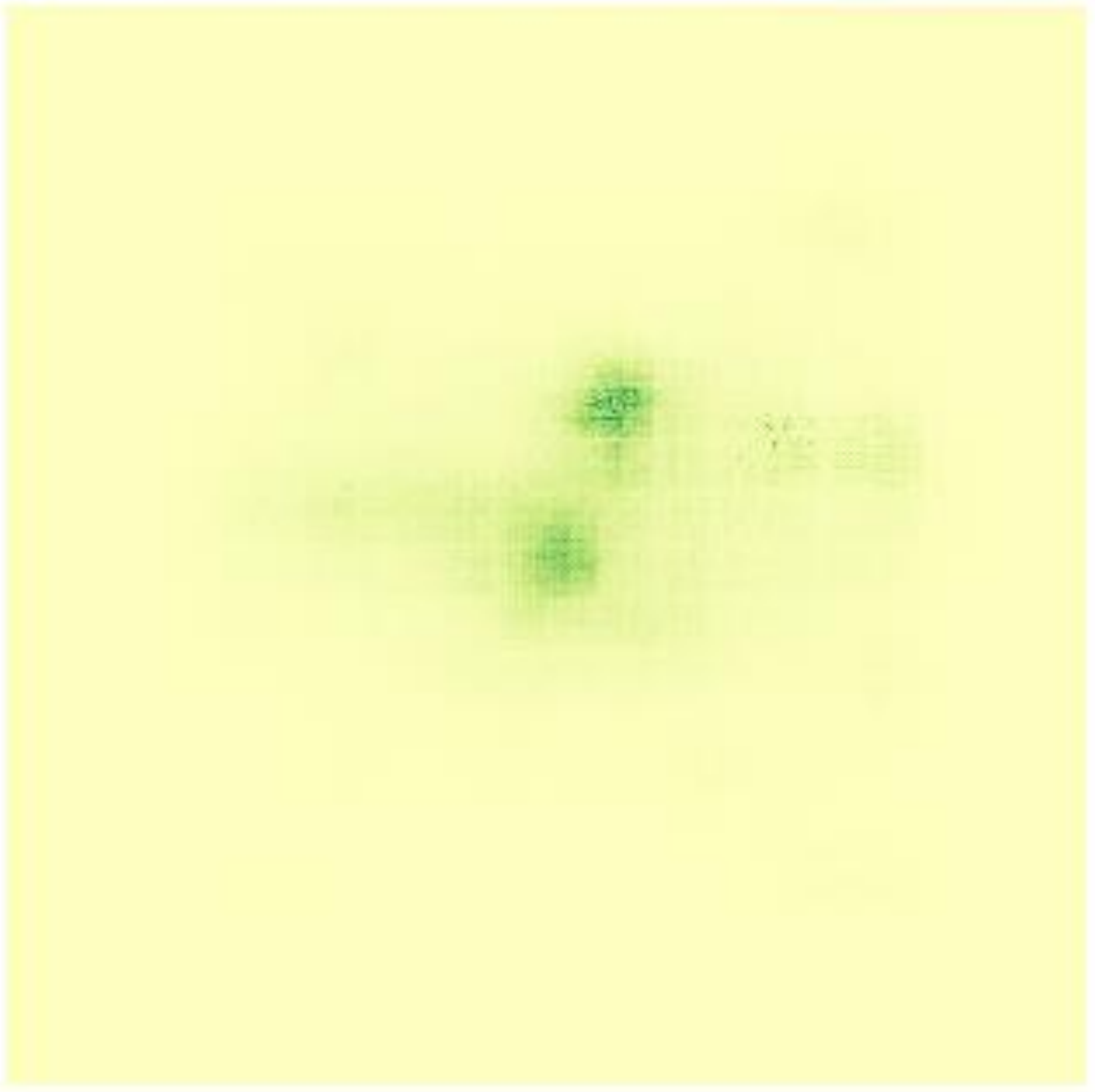}%
		\label{MS2Fusion_erf}}
	\caption{ Effective receptive field visualizations comparing CNN-based fusion method (a), Transformer-based fusion method (b), and the proposed MS2Fusion (c) method. Quantitative analysis demonstrates that MS2Fusion achieves significantly broader receptive field coverage compared to the others.}
	\label{erf_com}
\end{figure*}

Recent advances in sequence modeling show SSM-based methods excel by compressing features into compact hidden states, enabling efficient inference with constant-time full-sequence processing. Mamba \citep{gu2023mamba} enhances this with selective state spaces, dynamically retaining task-relevant features. Vision Mamba (Vim) \citep{zhu2024vision} further proves its effectiveness for visual tasks, boosting both efficiency and performance.

Inspired by this, we propose MS2Fusion, a novel framework that simultaneously leverages complementary and shared features across modalities while overcoming limitations of CNN and Transformer in multispectral feature fusion. \Cref{dif_compare} reveals that Transformer-based methods (a) and existing Mamba solutions (b) fail to exploit cross-modal shared features. In contrast, our method (c) explicitly models both complementary and shared cross-modal interactions through three key components:

\begin{itemize}
    \item Cross-Parametric State Space Model (CP-SSM): Facilitates implicit feature complementarity by exchanging output matrices between modality-specific state spaces, enabling cross-modal feature enrichment while preserving modality-specific characteristics.
    \item Shared-Parametric State Space Model (SP-SSM): Learns a unified feature space through parameter sharing, aligning heterogeneous modality distributions to extract discriminative shared representations that enhance single-modality features.
    \item Feature Fusion State Space Model (FF-SSM): Introduces a bidirectional input scheme to expand the Effective Receptive Field (ERF) of state spaces, mitigating feature attenuation while enabling adaptive fusion of cross-modal information.
\end{itemize}

As demonstrated in \Cref{erf_com}, our ERF analysis reveals that MS2Fusion achieves superior spatial coverage compared to both CNN and Transformer, successfully integrating local details with global context. 
This capability addresses the fundamental constraint of standard CNN in modeling long-range dependencies while maintaining computational efficiency.
Table \ref{flops} also provides our component replacement experiments on the MS2Fusion module in the YoloV5 framework on FLIR dataset(Figure \ref{overall architecture}), which yield three key findings:
(1) Compared to Transformer, MS2Fusion achieves a 66.6\% reduction in FLOPs and 70.4\% fewer parameters while delivering better detection accuracy (+8.3 mAP50 points);
(2) When benchmarked against CNN baseline, it maintains 26.0\% lower computational costs and 18.4\% parameter reduction, while achieving significant improvements of +9.0 mAP50 points;
(3) Across all complexity-accuracy trade-off metrics, MS2Fusion consistently outperforms both baseline models. Notably, our approach achieves processing speed comparable to CNN (with only a 2.1 frames difference) while being significantly faster than Transformer (by 10.5 frames).
These systematic experiments provide conclusive evidence that MS2Fusion successfully breaks the traditional efficiency-accuracy trade-off, establishing new state-of-the-art performance in multispectral object detection.

\begin{table*}[!t]
	\centering
	\footnotesize
	\caption{Comparison of different multispectral feature fusion methods(YoloV5 framework, FLIR dataset)}
	\begin{tabular}{ccccccccl}
		\toprule
		         & GFLOPs & Params (M) &FPS& mAP@0.5& mAP@0.75& mAP \\
		\midrule
		CNN          & 190.3 & 159.7 &30.8&74.3&23.8&32.5  \\
Transformer & 421.9 & 440.6 & 19.2&75.0&24.1&33.4    \\
MS2Fusion          & 140.8 & 130.3  &29.7& 83.3&33.0&40.3  \\
		\bottomrule
	\end{tabular}
    
	\label{flops}
\end{table*}

 Our contributions are as follows: 
\begin{itemize}
	\item[$\diamond$] To address the limitations of existing methods that either overlook cross-modal interactions or overly rely on complementary features, \\MS2Fusion introduces a novel state-space framework. Its CP-SSM captures implicit feature complementarity while SP-SSM enhances shared representations, achieving an optimal balance between modality-specific and cross-modal learning.
	\item[$\diamond$] Motivated by the need of combining global context modeling with computational efficiency, our FF-SSM mechanism uniquely achieves Transformer-level receptive fields while maintaining CNN-like efficiency. This breakthrough resolves the longstanding trade-off between performance and computational cost in multimodal systems.
	\item[$\diamond$] MS2Fusion architecture offers flexible multimodal integration, supporting various detection frameworks and backbones. It seamlessly processes RGB, thermal, and other modalities, enabling broad application across different vision tasks without architectural changes.
	\item[$\diamond$] The MS2Fusion achieves state-of-the-art performance on benchmark datasets, including RGB-T object detection, semantic segmentation, and salient object detection, validating its effectiveness for multispectral image perception.
\end{itemize}

\par The rest of this paper is organized as follows. \Cref{sec2} reviews related research on multi-spectral object detection and Mamba. \Cref{sec3} describes the proposed method. \Cref{sec4} presents experimental results and analysis. Finally, we summarize the main points of the paper in \Cref{sec5}.

\section{Related Works}
\label{sec2}
\subsection{Object Detection}
\label{subsec2.1}

In object detection, RGB images are typically used for unimodal detection. There are two main approaches: two-stage detectors and one-stage detectors. Two-stage detectors (e.g., R-CNN \citep{girshick2014rich}, Fast R-CNN \citep{girshick2015fast}, Faster R-CNN \citep{ren2015faster}, Mask R-CNN \citep{he2017mask}) first generate region proposals and then perform classification and bounding box regression. 
Single-stage detectors (e.g., YOLO \citep{redmon2016you}, SSD \citep{liu2016ssd}, RetinaNet \citep{lin2017focal}) perform object detection directly on the image without generating region proposals, resulting in faster detection speeds. 
\par
Anchor-based methods in object detection utilize predefined anchor points, each representing specific sizes and aspect ratios, to detect objects through regression adjustments. 
In contrast, anchor-free methods such as CornerNet \citep{duan2019centernet}, FCOS \citep{9010746}, and CenterNet \citep{9010985} predict object boundaries or centers directly without anchors, simplifying the detection process with improved efficiency and often higher accuracy. 
\par
More recently, DETR (DEtection TRansformer) \citep{carion2020end} introduced a fully end-to-end approach by leveraging Transformer to eliminate the need for hand-designed components like anchors or non-maximum suppression (NMS). DETR treats object detection as a set prediction problem, using bipartite matching to assign predictions to ground truth objects. While achieving competitive accuracy, its computational cost and slow convergence remain challenges, prompting follow-up improvements like Deformable DETR \citep{zhu2021deformable}.

To fully validate the effectiveness of the proposed method, we selected the anchor-based YOLOv5 detection framework and the Transformer-based CoDetr detection framework for comparative experiments.
The experimental results show that the proposed method offers the following advantages: 1) linear time complexity, ensuring high computational efficiency; 2) a global receptive field characteristic, enabling the capture of richer contextual information.

\subsection{Multispectral Object Detection}
\label{subsec2.2}
Recent advances in multispectral object detection have made significant progress in addressing two core challenges: cross-modal feature fusion and environmental adaptability. Early approaches focused on balanced feature integration, with methods like the Cyclic Fusion Module \citep{zhang2020multispectral} explicitly modeling both complementarity and consistency between modalities. Subsequent works introduced more sophisticated attention mechanisms to dynamically weight features, such as Guided Attention Feature Fusion (GAFF) \citep{zhang2021guided}, which employed adaptive intra-modal and cross-modal attention to enhance fusion performance.

The emergence of Transformer-based architectures has further advanced the field by capturing long-range dependencies between modalities. The Cross-Modal Fusion Transformer (CFT) \citep{qingyun2021cross} demonstrated the effectiveness of self-attention for global contextual fusion, while CMX \citep{zhang2023cmx} improved generalization through feature rectification modules. Recent variants like ICAFusion \citep{shen2024icafusion} and INSANet \citep{lee2024insanet} have optimized efficiency and flexibility, using parameter-shared Transformer and dedicated spectral attention blocks, respectively.

Several innovative approaches have pushed the boundaries of multispectral detection by addressing modality-specific challenges. DAMSDet \citep{guo2024damsdet} tackles modality misalignment and dynamic complementary characteristics through deformable cross-attention, while MS-DETR \citep{xing2023multispectral} introduces reference-constrained fusion to improve RGB-thermal alignment. Lightweight designs have also gained attention, such as the CPCF module \citep{hu2024rethinking}, which combines channel-wise and patch-wise cross-attention for efficient fusion. Meanwhile, TFDet \citep{zhang2024tfdet} employs a fusion-refinement paradigm with adaptive receptive fields to suppress false positives. Most recently, MMFN \citep{yang2024multidimensional} proposed a comprehensive hierarchical fusion framework, integrating local, global, and channel-level interactions for robust multispectral detection.

Our method employs a novel SSM for feature fusion, which is different from the former methods using CNN or Transformer for feature fusion. 
\subsection{Mamba}
\label{sec2.3}

Mamba \citep{gu2023mamba} is a selectively structured state-space model designed for effective long sequence modeling tasks. It overcomes the limitations of CNN by incorporating global perceptual fields and dynamic weighting, achieving advanced modeling capabilities akin to Transformer but without their typical quadratic complexity.
Building on Mamba's foundation, VMamba \citep{liu2024vmamba} is a visual state-space model that introduces the Cross Scan Module (CSM) to enhance scanning efficiency across dimensions, surpassing both CNN and ViTs in performance for computer vision tasks. Meanwhile, VM-UNet \citep{ruan2024vm} integrates Mamba into the UNet framework for medical segmentation, leveraging visual state-space blocks to capture extensive contextual information.
Additionally,  Mamba is applied to multimodal semantic segmentation \citep{wan2024sigma}, enhancing global receptive field coverage with linear complexity through a Siamese encoder and innovative fusion mechanisms.\par
Inspired by these advancements, we propose MS2Fusion based on Mamba dynamic state space. This approach effectively harnesses shared features and complementarities between modalities, enhancing object detection accuracy while reducing model complexity. 
\section{Methods}
\label{sec3}
\subsection{State Space Model (SSM)}
\label{subsec3.1}
SSM \citep{gu2023mamba} combines the advantages of RNNs and CNNs for efficient handling of long dependencies. It transforms the input sequence $x(t)$ into an intermediate state $h(t)$ through a state equation, then generates the output $y(t)$ via an output equation. SSM is typically represented as a linear ODE:
\begin{equation}
	\centering
\begin{aligned}
    h^{'}(t) &= \textbf{A} \cdot h(t) + \textbf{B} \cdot x(t) \\
    y(t) &= \textbf{C} \cdot h(t) + \textbf{D} \cdot x(t)
\end{aligned}
\label{SSM}
\end{equation}
To adapt SSM for deep learning, it is discretized as:
\begin{equation}
	\centering
\begin{aligned}
    h_{k} &= {\overline{\textbf{A}}} \cdot h_{k - 1} + {\overline{\textbf{B}}} \cdot x_{k} \\
    y_k &= {\overline{\textbf{C}}} \cdot h_{k} + \textbf{D} \cdot x_{k}  \\
    {\overline{A}} &= exp^{{{\Delta}}\textbf{A}}\\
        {\overline{B}} &={\left( {exp^{{{\Delta}}\textbf{A}} - \textbf{I}} \right)}/{\Delta}\textbf{A}\\
        {\overline{C}} &= \textbf{C}
\end{aligned}
\label{dis_SSM}  
\end{equation}
After discretization \citep{zhang2023effectively}, SSM uses convolution operations for parallel computation. To overcome the issue of fixed parameters, SSM introduces dynamic adjustments, allowing the model to flexibly handle complex data and improve long-sequence modeling.

\begin{figure*}[!t]
	\centering
	\includegraphics[width=0.9\linewidth]{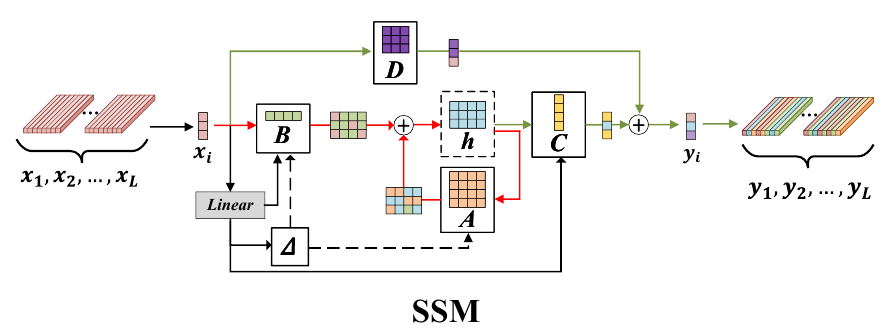}
	\caption{ Details of SSM, where the red, green and black lines correspond to the equations of \Cref{dis_SSM}, respectively. (For an $L \times d$ dimensional input \{$x_1, x_2, ...,x_L$\}.) The input sequence is linearly projected to generate parameters for the state equation, followed by recursive computation via matrices \textbf{A},\textbf{ B}, and \textbf{C}, with an optional skip connection \textbf{D}. The output sequence retains the same dimensionality as the input.}
	\label{SSM_detail}
\end{figure*}
As shown in \Cref{SSM_detail}, the SSM processes the input sequence (\(\{x_1, x_2, ..., x_L\} \in \mathbb{R}^{L \times d}\)) through a structured transformation composed of three key operations. First, the input undergoes a linear projection to generate parameters for the state space equation. This is followed by a recursive state update governed by four fundamental matrices: the current input is initially processed by matrix \textbf{B}, which maps the input to the state space, and then added to the hidden state updated by matrix \textbf{A}, producing a new hidden state. For the output, the new hidden state is remapped to the output via matrix \textbf{C}. Additionally, \(\textbf{D}\) provides an optional skip connection that bypasses the state transformation, directly adding the input to the output, which helps mitigate gradient vanishing during training. This recurrent computation generates intermediate outputs \(\{y_1, y_2, ..., y_L\} \in \mathbb{R}^{L \times d}\) \citep{zhang2023effectively}, where each \(y_i \in \mathbb{R}^{1 \times d}\) is derived sequentially. 
\par
\begin{figure*}[!t]
	\centering
	\includegraphics[width= 0.8\linewidth]{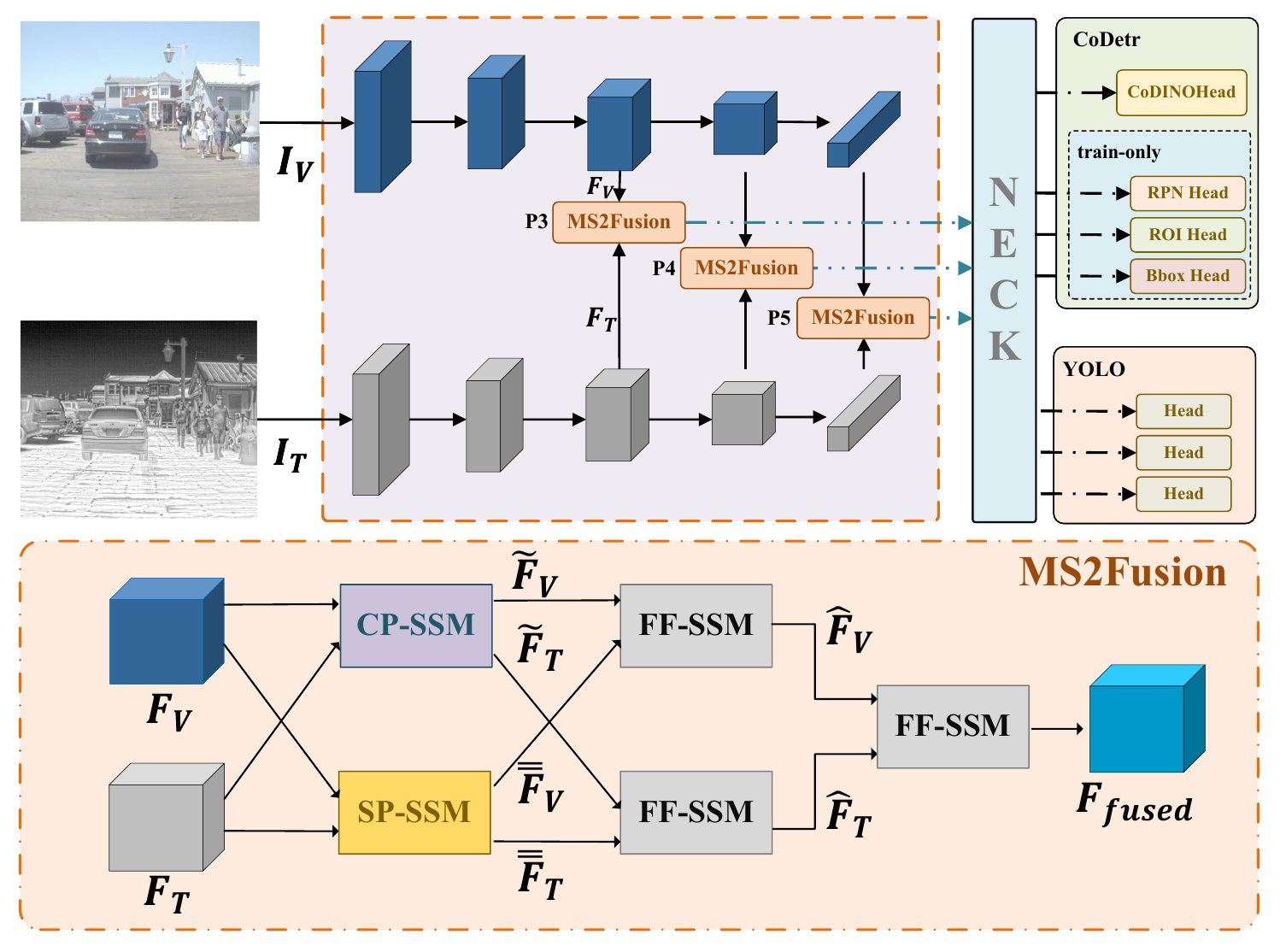}
	\caption{ Overview of the model architecture. It consists of three main stages: (1) feature extraction with two backbone networks; (2) cross-modal feature fusion of $P3$, $P4$ and $P5$ with MS2Fusion module; (3) the detection results are generated through the Neck and Head layers. In our experiments, two distinct detection heads ( CoDetr and YOLOv5) are evaluated independently. The MS2Fusion module employs a dual-branch architecture to process the features of two modalities, $F_V$ and $F_T$. CP-SSM learns cross-modal complementary features, while SP-SSM extracts shared features. Finally, FF-SSM performs single brach feature enhancement and cross-modal fusion, outputting the fused feature $F_{fused}$.}
	\label{architecture}
\end{figure*}

\subsection{The Proposed Model Architecture}
\label{subsec3.2}
As shown in the upper part of \Cref{architecture}, the proposed framework is structured into three main stages. Initially, the backbone network extracts features independently from RGB and thermal images. Subsequently, cross-modal MS2Fusion feature fusion integrates features from different stages. Finally, object localization and regression are performed using the detection head to derive the final detection outcomes.
During the feature fusion stage, we selectively merge features from three distinct levels. P3 layer captures detailed surface information, while P5 encapsulates higher-level semantic feature. By separately fusing shallow and deep features, our approach effectively concentrates on detailed information critical for cross-modal fusion, which is formulated in \Cref{overall architecture}:

\begin{equation}
    \begin{aligned}
        F_{fused}^{i} &= {\phi_{MS2Fusion}\left( {F^i_V, F^i_T} \right)}\\
		\left\lbrack {bbox,cls} \right\rbrack&= {\Phi_{head}\left( {\Psi_{neck}\left( F_{fused}^{3},F_{fused}^{4},F_{fused}^{5} \right)} \right)} 
    \end{aligned}
    \label{overall architecture}
\end{equation}
where $F^i_V, F^i_T$ denote the feature maps in layer $i$ extracted from the input RGB and thermal image with two backbones, and $\phi_{MS2Fusion}$ is the proposed state-space fusion module, $F_{fused}^i$ denotes the fused feature. FPN \citep{lin2017feature} and PANet \citep{liu2018path} are usually used as neck($\Psi_{neck}$) to aggregate multi-scale features, while the detection head($\Phi_{head}$) is used for bounding box classification and regression. In our experiments, two different detection heads (YOLOv5 and CoDetr) are evaluated separately.

\subsection{Multispectral State Space Feature Fusion (MS2Fusion)}
\label{subsec3.3}
As shown in the lower part of \Cref{architecture}, the MS2Fusion module consists of three core components: CP-SSM, SP-SSM, and FF-SSM. CP-SSM module achieves global feature integration of RGB and thermal modalities through a dynamic parameter interaction mechanism, enhancing cross-modal contextual awareness while preserving modality-specific characteristics. This module innovatively employs an implicit parameter crossover strategy to effectively mine and reinforce complementary features between the two modalities. CP-SSM allows adaptive parameter modulation based on inter-modal relationships, enabling finer-grained complementary feature extraction. In contrast, the SP-SSM is designed to learn shared, modality-invariant representations that are consistent across both RGB and thermal inputs. All parameters in SP-SSM are constrained to be identical between modalities, forcing the model to discard modality-specific noise and to focus on common semantic information, such as object contours, shapes, and scene structure. 

The key distinction between CP-SSM and SP-SSM lies in their parameter handling strategies and functional objectives. The CP-SSM employs dynamic parameter interaction mechanisms where modality-specific parameters are selectively exchanged or combined to capture complementary cross-modal dependencies, while the SP-SSM strictly maintains identical parameters across modalities to enforce feature consistency in shared spaces. 

Finally, the FF-SSM modules further integrate these learned features in a hierarchical manner, adaptively combining complementary and shared information to achieve more effective cross-modal fusion.

The innovation of MS2Fusion lies in its systematic utilization of the dual characteristics of multispectral data: the CP-SSM explores cross-modal complementary information, while the SP-SSM strengthens modality-invariant shared features, ultimately achieving optimal feature fusion through the FF-SSM. This dual-path design explicitly decouples the learning of complementary and shared features: CP-SSM’s parameters are optimized for cross-modal differentiation, whereas SP-SSM’s parameters are optimized for cross-modal unification. By simultaneously focusing on modality-specific complementary features and modality-invariant shared features, the framework establishes a comprehensive cross-modal feature collaboration mechanism that significantly enhances the representational capability of multispectral data.
\subsubsection{CP-SSM Module}
\label{subsubsec3.3.1}

\begin{figure*}[t]
	\centering
	\includegraphics[width=0.8\linewidth]{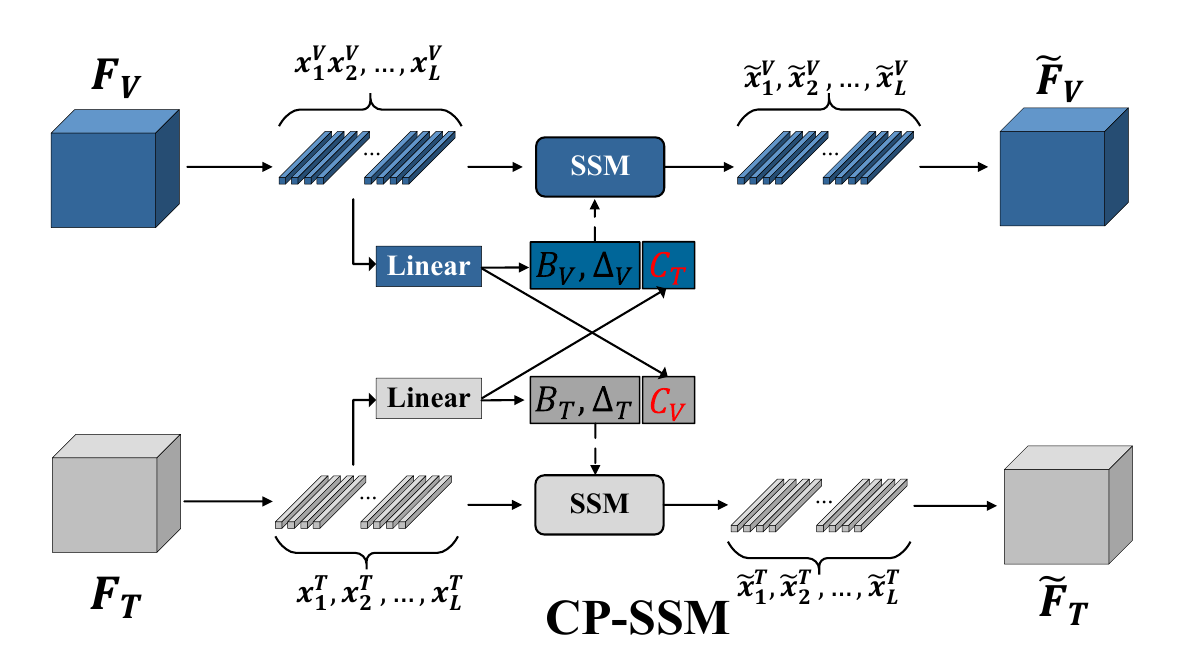}
	\caption{The details of the CP-SSM module. The feature maps ($F_V, F_T$) are first reshaped into a sequence ($x_i^V$, $x_i^T$)  by row and column scanning, and generated \textbf{$B, C, \Delta$} through a Linear layer. Secondly, we perform a cross-modal complementary features interaction by exchanging the C of the two branches. Finally, the cross-modal complementary feature interaction is conducted by the SSM module to generate the output ($\widetilde{F}_V, \widetilde{F}_T$).}
	\label{CP-SSM}
\end{figure*}

As illustrated in \Cref{CP-SSM}, there are two branches in the CP-SSM module, RGB feature (top) and thermal feature branch (bottom). In this section, we only describe the RGB feature branch for clarity. The procedure of the thermal feature branch is identical.

Given $F_V\in \mathbb{R}^{d\times H\times W}$ from the RGB feature map, it is unfolded into a sequence $(x_1,x_2,x_3,\ldots,x_L)\in \mathbb{R}^{d\times L}$, where $d$, $H, W$ denote the channel, height and width of the feature map, respectively and $L=H \times W$. Following the Mamba mechanism in \Cref{subsec3.1}, the unfolded sequences are passed through a linear layer to obtain $\textbf{B}, \textbf{C}, \Delta,$ where $B\in \mathbb{R}^{L\times d\times d'}, C\in \mathbb{R}^{L\times d\times d'},\Delta \in \mathbb{R}^{L \times d}$, and $d'$ is the dimension of the hidden state. 
In the CP-SSM module, we innovatively designed a cross-parameter interaction mechanism that achieves implicit feature fusion through real-time interaction between the dual-modal state space projection matrices $C_V$ and $C_T$. Specifically, this mechanism establishes bidirectional feature enhancement channels: in the RGB modality branch, the response patterns of thermal features are selectively fused via the exchanged $C_T$ matrix, while the thermal branch enhances its feature discriminability by incorporating the semantic prior information encoded in the $C_V$ matrix. This cross-parameter interaction strategy possesses two key characteristics: (1) it maintains the independence of modality-specific features to avoid confusion, and (2) it constructs an implicit attention mechanism at the state space dimension, enabling the complementary feature fusion process to be self-adaptive. Finally, after the reverse operation of folding the sequence $(\Tilde{x}_1,\Tilde{x}_2,\Tilde{x}_3,\ldots,\Tilde{x}_L)$, the output $\Tilde{F}\in \mathbb{R}^{d\times H\times W}$ for each branch is obtained in \Cref{CP-SSM_eq}:
\begin{equation}
    \centering
	\begin{aligned}
		\Tilde{F}_{i} &= \varphi_{SSM}^{i}\left( {F_{flip}^{i}\left( F_{i} \right),\Delta_{i},B_{i},C_{i}} \right) \\
        \left\lbrack {\Delta_{i},B_{i},C_{i}} \right\rbrack &= L_{Linear}\left( {F_{i}} \right),  i \in \{V,T\}
	\end{aligned}
	\label{CP-SSM_eq}
\end{equation}
where $F_{flip}^{i}$ indicates that the feature map is expanded in a certain way into a sequence, $B_i$, $C_i$, $\Delta_i$ obtained from the input sequence through the fully connected layer, and $\varphi_{SSM}^{i}$ denotes the SSM in \Cref{SSM_detail}.

It is worth noting that, inspired by the VSSBlock in VMamba \citep{liu2024vmamba}, we have explored three unfolding methods: unfold by rows, unfold by columns and unfold with both. We will provide detailed ablation studies in \Cref{finetune}.

\begin{figure*}[t]
	\centering
	\includegraphics[width=0.8\linewidth]{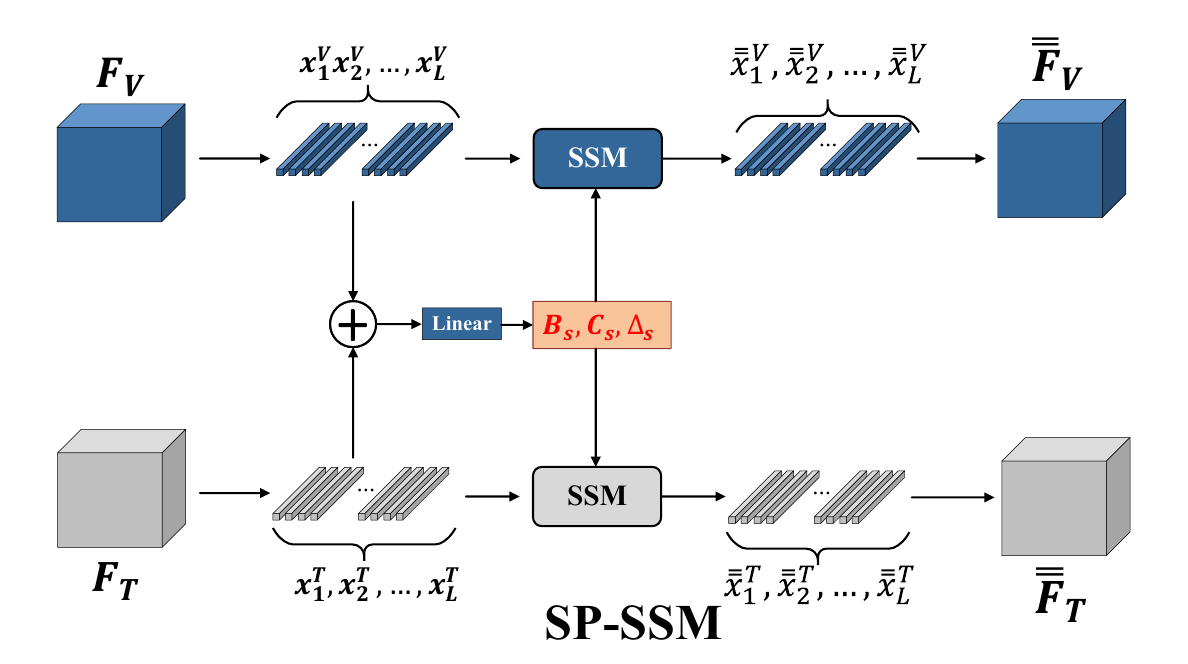}
	
	\caption{The details of the SP-SSM module. The SP-SSM module extracts shared features in two modalities (RGB and thermal) from the SSM with shared parameters. The input feature  $F_V$ and $F_T$ are combined to generate parameter \textbf{$B_s, C_s, \Delta_s$}, while the output feature $\overline{\overline{F}}_V$ and $\overline{\overline{F}}_T$ are reconstructed by the two SSMs.
	}
	\label{SP-SSM_detail}
\end{figure*}

\begin{figure*}[t]
	\centering
	\includegraphics[width=0.8\linewidth]{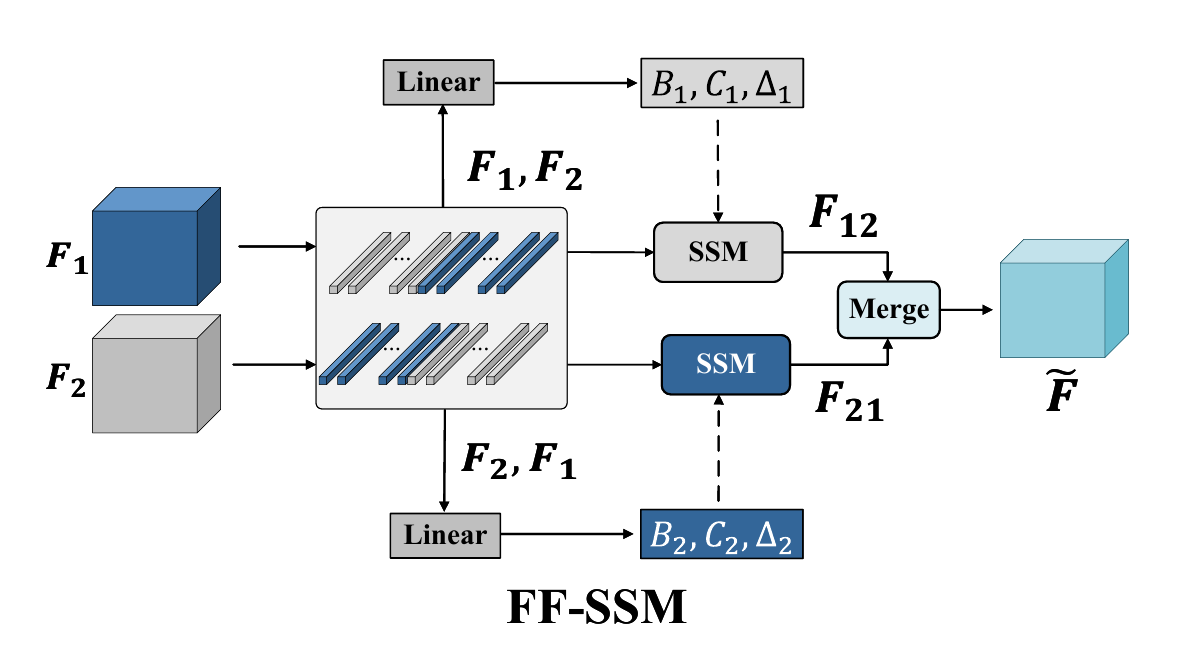}
	\caption{The details of the FF-SSM module. It fuses features by combining $F_1$ and $F_2$ in two different orders. In the top path, the input features are combined in 1-2 order (e.g., $F_1, F_2$) to form the splice feature to generate \textbf{$B_1, C_1, \Delta_1$} by a Linear layer, and then cross feature interactions are performed via SSM. Finally, the $F_{12}$ and $F_{21}$ are merged to generate the fused feature map $\widetilde{F}$.}
	\label{FF-SSM_detail}
\end{figure*}
\subsubsection{SP-SSM Module}
\label{subsubsec3.3.2}

The SP-SSM module constructs a hierarchical feature-sharing architecture, achieving cross-modal representation alignment through parameter sharing and feature reconstruction. As shown in \Cref{SP-SSM_detail}, the pipeline of this module consists of two meticulously designed stages:

In the parameter-sharing stage, the module employs a coarse-grained additive feature fusion method to preliminarily integrate the RGB and thermal features. 
Then, three sets of critical shared parameters are dynamically generated through a linear network. These parameters not only encode the common feature patterns of the dual modalities but also retain the modality-specific adjustment capabilities.

In the feature reconstruction stage, these shared parameters are injected into the SSM of both modalities. 
This design creates a dual-stream coupled architecture: on one hand, the shared parameters constrain the evolution trajectories of the two modalities in the state space, driving heterogeneous features to converge into a shared feature space; on the other hand, each modality retains independent initialization states, ensuring the integrity of modality-specific information. 
Through this shared-parameter computation paradigm, the module can extract deep shared features that are invariant to factors such as lighting conditions and environmental interference.
The specific process is formulated in \Cref{SP-SSM}:
\begin{equation}
	\label{SP-SSM}
	\begin{aligned}
		\overline{\overline{F}}_i &= \varphi_{SSM}\left( {F_{i}, \Delta_{s},B_{s},C_{s}} \right) \\
		\left\lbrack {\Delta_{s},B_{s},C_{s}} \right\rbrack &= L_{Linear}\left( {F_{V} \oplus F_{T}} \right)
	\end{aligned} 
\end{equation}
where $\overline{\overline{F_i}}$ $(i\in\{V, T\})$ are the shared features and $\varphi_{SSM}$ is the SSM. $\Delta_{s}, B_{s}, C_{s}$ are obtained from the weighted fusion features through a fully connected layer.
\subsubsection{FF-SSM Module}
\label{subsubsec3.3.3}

Although the CP-SSM module can capture implicit complementary relationships between modalities, it still has limitations at the feature fusion level. To address Mamba's inherent feature forgetting while preserving its computational efficiency, we propose a bidirectional architecture combining complementary forward ($F_1, F_2$) and reverse ($F_2, F_1$) processing paths. This innovative design systematically compensates for unidirectional modeling limitations through direction-adaptive parameter generation, where the reverse path specifically aims to capture potential feature loss in the forward pass. Unlike traditional Transformers that suffer from information loss due to chunking operations, our approach not only maintains Mamba's original sequential dependency and linear complexity advantages but also significantly enhances feature retention through reverse-sequence compensation. The dual-path mechanism enables more comprehensive capture of long-range cross-modal dependencies while dynamically mitigating feature forgetting, achieving robust sequential modeling without sacrificing efficiency. Specifically, given the feature maps $F_1$ and $F_2$ output by CP-SSM, this module employs a bidirectional heterogeneous sequence construction strategy: concatenating the features in different orders ($F_1, F_2$ and $F_2, F_1$). This design not only expands the model's receptive field but also enhances the diversity of feature representation. In implementation, the features are first unfolded and concatenated for the two directional sequences, followed by generating corresponding state space model parameters ($B_1, C_1, \Delta_1$ and $B_2, C_2, \Delta_2$) through linear transformation layers. Finally, by fusing the output features ($F_{12}$ and $F_{21}$) from the bidirectional SSM paths, the module achieves thorough cross-modal feature interaction and adaptive fusion, significantly improving detection accuracy.
The FF-SSM module effectively preserves the detailed information within both modalities, ensuring richer and more precise feature representation, which is formulated in \Cref{FF-SSM}:
\begin{equation}
\label{FF-SSM}
\begin{aligned}
    \widehat{F} &= \psi_{Merge}\left(F_{12}, F_{21} \right) \\
		F_{12} &= {\varphi_{SSM}\left( {\left\lbrack {F_{1},F_{2}} \right\rbrack,\Delta_{1},B_{1},C_{1}} \right)} \\
		F_{21} &= {\varphi_{SSM}\left(\left\lbrack {F_{2},F_{1}} \right\rbrack,\Delta_{2},B_{2},C_{2}\right)} \\
		\left\lbrack {\Delta_{1},B_{1},C_{1}} \right\rbrack &= L_{Linear}^{1}\left( {F_{1},F_{2}} \right) \\
		\left\lbrack {\Delta_{2},B_{2},C_{2}} \right\rbrack &= L_{Linear}^{2}\left( {F_{2},F_{1}} \right)
\end{aligned}
\end{equation}
\noindent where $\varphi_{SSM}$ denotes the SSM in \Cref{SSM_detail}, $F_1, F_2$ denotes the input feature maps from RGB and thermal modalities, $\psi_{Merge}$ denotes the merging of the outputs of two sequences with different connection orders after the SSM. The  $\Delta_i$, $B_i$, $C_i$ are the corresponding parameters in SSM.
\subsection{Loss Function}
\label{subsec3.4}
The fused features produced by MS2Fusion are fed into the neck and detection head, with the entire pipeline being optimized end-to-end.
In this paper, we have evaluated MS2Fusion in both YOLOv5 and Co-Detr based detection frameworks. 

In the YOLO framework, the total loss function can be formulated as \Cref{loss}:
\begin{equation}
	\centering
	\begin{aligned}
		\mathcal{L}_{yolo}  = \lambda_{\text {bbox }}\cdot \mathcal{L}_{\text {bbox}}+ \lambda_{\text {obj }}\cdot\mathcal{L}_{\text {obj}}+\lambda_{\text {cls }}\cdot\mathcal{L}_{\text {cls}}	
	\end{aligned}
	\label{loss}
\end{equation}
where $\mathcal{L}_{\text {bbox}}$, $\mathcal{L}_{\text {obj}}$ and $\mathcal{L}_{\text {cls}}$ are the localization loss,  confidence loss and classification loss, respectively. The hyperparameters $\lambda_{bbox}$, $\lambda_{obj}$ and $\lambda_{cls}$  adjust the loss weights, which are kept as the default settings of the baseline.\par

The CoDetr framework employs a multi-task loss function consisting of four components: a primary CoDINOHead~\citep{zhou2024optimizing} and three auxiliary detection heads (RPN head, ROI head, and Bbox head).
The total loss can be formulated in \Cref{codetr_loss}: 
\begin{equation}
	\centering
\begin{aligned}
\mathcal{L}_{\text{CoDetr}} &= \lambda_{\text{primary}}\cdot(\mathcal{L}_{\text{QFL}} + \mathcal{L}_{\text{L1}} + \mathcal{L}_{\text{GIoU}}) + \lambda_{\text{RPN}}\cdot(\mathcal{L}_{\text{CE}} + \mathcal{L}_{\text{L1}}) \\
&+ \lambda_{\text{ROI}}\cdot(\mathcal{L}_{\text{CE}} + \mathcal{L}_{\text{GIoU}}) + \lambda_{\text{Bbox}}\cdot(\mathcal{L}_{\text{Focal}} + \mathcal{L}_{\text{GIoU}} + \mathcal{L}_{\text{CE}})
\end{aligned}
\label{codetr_loss}
\end{equation}
where $\mathcal{L}_{\text{QFL}}$, $\mathcal{L}_{\text{L1}}$, $\mathcal{L}_{\text{GIoU}}$, $\mathcal{L}_{\text{CE}}$ and $\mathcal{L}_{\text{Focal}}$ are the quality focal loss, L1 loss, GIoU loss, cross-entropy loss and focal loss. Hyperparameters $\lambda$s are the weighting factor for each loss term.

\section{Experiments}
\label{sec4}
\subsection{Dataset and Evaluation Metric}
\label{subsec4.1}
\textbf{FLIR} \citep{FLIR}: It contains 5,142 RGB-T image pairs captured during both daytime and nighttime.
Due to the misalignment in the original dataset, the aligned version \citep{2020Multispectral} is commonly chosen for the experiments. The dataset is divided into 4,129 pairs for training and 1,013 pairs for testing.\par

\textbf{LLVIP} \citep{jia2021llvip}: It contains street scenes rich in pedestrians and cyclists, amounting to 15,488 RGB-T image pairs \citep{jia2021llvip}. Following~\citep{jia2021llvip}, 12,025 pairs are used for training and 3,463 pairs for testing. 
Thermal images are primarily used for labeling, which is copied directly to RGB images.\par

\textbf{M$^3$FD} \citep{liu2022target}: It includes 4,200 RGB-T aligned image pairs collected under various conditions such as different lighting, seasons, and weather scenarios. It covers six typical categories of automated driving and road surveillance, which is divided into training and testing sets with a ratio of 8:2 as provided in \citep{Liang2023explicit}.\par
\textbf{VEDAI \cite{Jurie2016Vehicle}:} It is designed for multi-class vehicle detection in aerial images, comprising 3,640 objects instances across nine categories, including boats, cars, campers, airplanes, shuttles, tractors, trucks, vans, and others. It contains a total of 1,210 images with a resolution of 1,024 × 1,024, each featuring three RGB color channels and one NIR channel. The dataset is divided into training and test sets in a 9:1 ratio. Annotations, originally in a rotated box format with quadratic coordinates, are converted to a horizontal-box format following the methodology described in \cite{QINGYUN2022108786}.\par
\textbf{Average Precision (AP): }The AP metric is derived from the area under the Precision-Recall curve, which plots recall on the horizontal axis and precision on the vertical axis. 
The mean Average Precision (mAP) is calculated by taking the weighted average of AP across all classes. In our experiments, we use an Intersection over Union (IoU) threshold of 0.5 to compute the mAP. Higher values of this metric indicate better performance. 
\subsection{Experimental Setup}
\label{subsec4.2}
All experiments are conducted using PyTorch on a computer equipped with an Intel i7-9700 CPU, 64 GB RAM, and a Nvidia RTX 3090 GPU with 24 GB of memory. For all ablation studies, the number of epochs is set to 60. 
In our experiments, the batch size is set to 4, and the SGD optimizer is used with an initial learning rate of 0.01 and a momentum of 0.937. The weight decay factor is set to 0.0005, and we employ a cosine learning rate decay schedule. The input size for all training images is $640 \times 640$, while the input size for testing is $640 \times 512$. Additionally, mosaic augmentation is used for data enhancement.

\subsection{State-of-the-art Comparison}
\label{subsec4.3}
The MS2Fusion module is experimented with  both the YOLOv5 and CoDetr framework \citep{zhou2024optimizing}. 
The YOLOv5 detector possesses faster inference speed but lower accuracy, while the CoDetr has better detection accuracy but slower inference speed. 
\subsubsection{Comparison on the FLIR Dataset}
\label{subsec4.3.1}
\begin{table*}[!t]
	\centering
	\footnotesize
	\caption{Comparison on the FLIR-align dataset. ('-' indicates missing values. '$\ddag$' symbol denotes experimental results obtained using the CoDetr framework, with input images resized to a fixed resolution of 640×640 pixels.)}
    \resizebox{!}{!}{
	\begin{tabular}{rccccc}
		\toprule
	Methods & mAP@0.5 &mAP&  Bicycle & Car & Person\\
	\midrule
	MMTOD-CG \citep{devaguptapu2019borrow} &	61.4&-&	50.3&	70.6&	63.3\\
	MMTOD-UNIT \citep{devaguptapu2019borrow} &	61.5&-&	49.4&	70.7&	64.5\\
	CMPD \citep{li2022confidence} &	69.4&-&	59.9&	78.1&	69.6\\
	CFR \citep{zhang2020multispectral} &	72.4&-&	57.8&	84.9&	74.5\\
	GAFF \citep{zhang2021guided} &	72.9&37.5&	-&	-&	-\\
	BU-ATT \citep{kieu2021bottom} &	73.1&-&	56.1&	87.0&	76.1\\
	BU-LTT \citep{kieu2021bottom} &73.2&-	&	57.4&	86.5&	75.6\\
	UA\_CMDet \citep{sun2022drone} &	78.6&-&	64.3&	88.4&	83.2\\
	CFT \citep{qingyun2021cross} &	78.7&40.2&	-&	-&	-\\
	CSAA \citep{cao2023multimodal} &	79.2&41.3&	-&	-&	-\\
	ICAFusion \citep{shen2024icafusion} &79.2&41.4&	66.9&	89.0&  81.6\\
	CrossFormer \citep{lee2024crossformer} &	79.3&42.1&	-&	-&	-\\
	MFPT \citep{zhu2023multi} &	80.0&-&	67.7&	89.0&	83.2\\
        MMFN \citep{yang2024multidimensional}&   80.8&   41.7&65.5&91.2&85.7\\
	RSDet  \citep{zhao2024removal} &	81.1&41.4&	-&	-&	-\\
        MiPa \citep{medeiros2024mipa} &81.3&44.8&-&-&-\\
        UniRGB-IR \citep{yuan2024unirgb} &81.4&44.1&-&-&-\\
        CPCF \citep{hu2024rethinking}&   82.1&   44.6&   -&  -&  -\\
        GM-DETR \citep{liu2024gm} &  83.9 & 45.8 &- &- &- \\
        Fusion-Mamba(yolov5) \citep{dong2024fusion} & 84.3 & 44.4 & - &- &-\\
        Fusion-Mamba(yolov8) \citep{dong2024fusion} & 84.9 & 47.0 & - &- &-\\
        TFDet \citep{zhang2024tfdet} &86.6   &46.6&  -&  -&  -\\
        DAMSDet\citep{guo2024damsdet}&86.6&  49.3&-&-&-\\
	\midrule
	Ours &	83.3&40.3&	74.9&	89.8&	85.1\\
   Ours\ddag &	\textbf{87.8}&\textbf{49.7}&	\textbf{79.6}&	\textbf{93.4}&	\textbf{90.2}\\
	\bottomrule
	\end{tabular}}
	\label{FLIR_com}
\end{table*}

\begin{table}[!t]
	\footnotesize
	\centering
	\caption{Comparison on the LLVIP dataset.}
	\begin{tabular}{rcc}
		\toprule
		Methods	&mAP@0.5&	mAP\\
		\midrule
		DIVFusion \citep{tang2023divfusion}&	89.8&	52.0\\
		GAFF \citep{zhang2021guided}&	94.0&	55.8\\
		CSAA \citep{cao2023multimodal}&	94.3&	41.3\\
		ECISNet \citep{an2022effectiveness} &	95.7&	-\\
		RSDet \citep{zhao2024removal}&	95.8&	61.3\\
        UniRGB-IR \citep{yuan2024unirgb} &96.1&63.2\\
		UA\_CMDet \citep{sun2022drone} &	96.3&	-\\
        CPCF \citep{hu2024rethinking}&   96.4&   65.0\\
         Fusion-Mamba(yolov5) \citep{dong2024fusion}& 96.8 & 62.8\\
        Fusion-Mamba(yolov8) \citep{dong2024fusion}& 97.0 & 64.3\\
     MMFN \citep{yang2024multidimensional}&   97.2&   -  \\
		GM-DETR \citep{liu2024gm} &    97.4&   70.2\\
        TFDet \citep{zhang2024tfdet}&    97.9&   \textbf{71.1}    \\
        DAMSDet\citep{guo2024damsdet}&97.9   &69.6   \\
        MiPa \citep{medeiros2024mipa} &98.2&66.5\\
		\midrule
		Ours&	97.5 &65.5	 \\
        Ours\ddag &	\textbf{98.4} &70.6	 \\
		\bottomrule
	\end{tabular}
	\label{LLVIP_com}
\end{table}

\Cref{FLIR_com} provides a comparative analysis of our method against existing approaches on the FLIR-align dataset. The results demonstrate that our method achieves the highest mAP@0.5 score of 83.3\% in YOLOv5 framework and 87.8\% in CoDetr framework across all classes, marking a significant improvement of 2.2\% (YOLOv5) and 6.7\% (CoDetr) over current state-of-the-art methods. Notably, the performance gain is particularly pronounced in the 'person' and 'bicycle' classes, highlighting our fusion method's superior effectiveness in addressing challenges related to non-thermal and small objects.\par

\subsubsection{Comparison on the LLVIP Dataset}
\label{subsec4.3.2}
\Cref{LLVIP_com} shows the performance metrics of our model on the LLVIP dataset, where our approach achieves superior results in both mAP@0.5 and mAP compared to existing models. 
Specifically, our method can obtain 97.7\%(YOLOv5) and 98.4\%(CoDetr) in terms of mAP@0.5, outperforming conventional CNN and Transformer-based approaches. This result underscores the effectiveness of our model in achieving state-of-the-art performance on the LLVIP dataset.

\begin{table*}[htp]
	\centering
	\footnotesize
	\caption{Comparison on the M$^3$FD dataset.}
    \resizebox{\linewidth}{!}{
		\begin{tabular}{rcccccccc}
			\toprule
			Methods&	mAP@0.5&	mAP&	People&	Bus&	Car&	Motorcycle&	Lamp&	Truck\\
			\midrule
			DIDFuse \citep{li2021didfuse}&	79.0&	52.6&	79.6&	79.7&	92.5&	68.7&	84.7&	68.8\\
			SDNet \citep{zhang2021sdnet}&	79.0&	52.9&	79.4&	81.4&	92.3&	67.4&	84.1&	69.3\\
			RFNet \citep{xu2022rfnet}&	79.4&	53.2&	79.4&	78.2&	91.1&	72.8&	85.0&	69.0\\
			ReC \citep{2022ReCoNet} \citep{zhao2023cddfuse}&	79.5&	-&	79.4&	78.9&	91.8&	69.3&	87.4&	70.0\\
			U2F \citep{9151265} \citep{zhao2023cddfuse}&	79.6&	-&	80.7&	79.2&	92.3&	66.8&	87.6&	71.4\\
			DAMSDet \citep{guo2024damsdet}&   80.2&   52.9&-&-&-&-&-&-\\
			TarDAL \citep{liu2022target}&	80.5&	54.1&	81.5&	81.3&	\textbf{94.8}&	69.3&	87.1&	68.7\\
			DeFusion \citep{sun2022detfusion}&	80.8&	53.8&	80.8&	83.0&	92.5&	69.4&	87.8&	71.4\\
			CDDFusion \citep{zhao2023cddfuse}&	81.1&	54.3&	81.6&	82.6&	92.5&	71.6&	86.9&	71.5\\
			IGNet \citep{li2023learning}&	81.5&	54.5&	81.6&	82.4&	92.8&	73.0&	86.9&	72.1\\
			SuperFusion \citep{tang2022superfusion}&	83.5&	56.0&	83.7&	93.2&	91.0&	77.4&	70.0&	85.8\\
             Fusion-Mamba(yolov5) \citep{dong2024fusion}& 85.0 & 57.5& 80.3&92.8&91.9&73.0&84.8&87.1\\
       
             MMFN \citep{yang2024multidimensional}&   86.2&   -& 83.0&   92.1&   93.2&   73.7&   87.6&   87.4    \\
              Fusion-Mamba(yolov8) \citep{dong2024fusion} & 88.0 &61.9&84.3&94.2&92.9&80.5&87.5&88.8\\
			\midrule
			Ours&	89.4&59.7&85.6 &93.7 &93.9 &82.4 &\textbf{90.8} &89.9	 \\			
            Ours\ddag &\textbf{91.4}&\textbf{65.6}&\textbf{89.8} &\textbf{95.0} &94.7 &\textbf{88.2} &88.3 &\textbf{92.4} \\
            \bottomrule
	\end{tabular}}
	\label{M$^3$FD_com}
\end{table*}

\begin{table*}[htp]
	\centering
	\footnotesize
	\caption{Comparison on the VEDAI dataset.}
	\scalebox{0.8}{
		\begin{tabular}{rcccccccccc}
			\toprule
				Methods&mAP@0.5&	Car&	Truck&	Pickup&	Tractor&	Camper&	Ship&	Van&	Plane\\
			\midrule
			ICAFusion \cite{shen2024icafusion}&	76.6&		88.0&	67.5&	80.9&	74.9&	78.3&	56.9&	69.8&	96.7\\
			MidFusion1 \cite{QINGYUN2022108786} &	77.4&		88.2&	\textbf{84.9}&	81.7&	68.4&	72.2&	71.0&	53.6&	\textbf{99.5}\\
			MidFusion2 \cite{QINGYUN2022108786} &	78.9&	89.2&	78.8&87.6&	66.2&	73.2&	62.5&	73.9&	\textbf{99.5}\\
			Input Fusion1 \cite{QINGYUN2022108786}&	80.0&	88.7&	77.7&	79.5&	72.3&	75.7&	\textbf{78.6}&	72.1&	95.7\\
			Input Fusion2 \cite{QINGYUN2022108786} &	80.5&	89.1&	\textbf{84.9}&	78.7&	84.2&	70.1&	69.9&	68.8&	98.4\\
			YOLOFusion \cite{QINGYUN2022108786}&	81.6&	91.7&	78.1&	85.9&	71.9&	78.9&	71.7&	75.2&	\textbf{99.5}\\
			\midrule
			Ours&80.2&89.5&75.6&84.2&81.6&75.9&61.8&76.4&97.0\\
			Ours\ddag &\textbf{84.2} &	\textbf{93.3 }&78.9&\textbf{88.5 }&	\textbf{85.8} &	\textbf{80.6} &	65.5& \textbf{83.6} &97.1\\
			\bottomrule
	\end{tabular}}
	\label{VEDAI_com}
\end{table*}

\subsubsection{Comparison on the M\texorpdfstring{$^{3}$}{}FD Dataset}
\label{subsubsec4.3.3}
Table \ref{M$^3$FD_com} provides a comparative analysis of our model against existing methods on the M$^3$FD dataset. Our approach shows superior performance with a notable 3.2\%(YOLOv5) and 5.2\%(CoDetr) improvement over other models. This improvement is particularly significant for high heat source objects such as 'motorcycles', showcasing the efficacy of our fusion approach in effectively integrating features from thermal images.\par
\subsubsection{Comparison on the VEDAI dataset.}
\label{subsubsec4.3.4}
Table \ref{VEDAI_com} highlights that our method achieves state-of-the-art performance on the VEDAI dataset, even without specialized techniques for detecting small objects. This underscores our model's strength in focusing on the multi-modal features of small objects, proving its effectiveness. These results indicate that our approach excels in UAV image applications, ensuring robust performance in detecting small objects.
\subsection{Generalization to Other Multimodal Tasks} \label{subsec4.4}
\subsubsection{Experiments on RGB-T Semantic Segmentation}\label{subsec4.4.1} 
\textbf{Evaluation metrics:} Mean Intersection over Union (mIoU) is a commonly used evaluation metric for semantic segmentation models, which measures the model performance by calculating the ratio between the intersection and union of predictions and ground truth segments. 

\textbf{MFNet dataset \citep{ha2017mfnet}}: It is the first publicly available RGB-T dataset featuring pixel-level annotations. It consists of 1,569 aligned RGB-T image pairs captured in urban environments, with semantic labels for eight common driving-scene obstacles: cars, pedestrians, bicycles, curves, bus stops, guardrails, traffic cones, and speed bumps.

\textbf{SemanticRT \citep{ji2023semanticrt} dataset}: It consists of 11,371 high-quality, pixel-level annotated RGB-T image pairs. It is seven times larger than the existing MFNet dataset and covers a wide range of challenging scenes under unfavorable lighting conditions such as low light and pitch black. \par

\textbf{Comparison on the MFNet dataset.} 
As illustrated in Table \ref{MFNet_com}, our MS2Fusion method performs best in this dataset, significantly improving mIoU by 2.8\% compared to the baseline \citep{ha2017mfnet} (without any specialized design). 
The improvement is particularly noticeable on the `Bump' and `Curve' categories.
Our analysis reveals that the "Bump" and "Curve" categories primarily rely on local geometric features of object surfaces, which constitute the shared features across multi-modal data. MS2Fusion can effectively leverage these shared features to significantly improve segmentation performance for these two categories. 

\begin{table*}[htp]
	\centering
	\footnotesize
	\caption{Comparison on the MFNet dataset.}
	\resizebox{\linewidth}{!}{
		\begin{tabular}{rcccccccccc}
			\toprule
			Methods&mIoU&Car&Person&	Bike&	Curve&	Car Stop&Guardrail&Color Cone& Bump\\
			\midrule
			PSTNet \citep{shivakumar2020pst900}&  48.4&76.8& 52.6& 55.3& 29.6& 25.1& 15.1& 39.4& 45.0\\
			RTFNet \citep{sun2019rtfnet} &53.2& 87.4 &70.3& 62.7& 45.3& 29.8& 0.0& 29.1& 55.7\\
			FuseSeg \citep{sun2020fuseseg}& 54.5&87.9 &71.7& 64.6 &44.8& 22.7& 6.4& 46.9& 47.9\\
			AFNet \citep{xu2021attention}&54.6& 86.0& 67.4& 62.0 &43.0& 28.9& 4.6& 44.9& 56.6\\
			ABMDRNet \citep{zhang2021abmdrnet}&54.8& 84.8& 69.6 &60.3 &45.1& 33.1& 5.1& 47.4& 50.0\\
			FEANet \citep{deng2021feanet} &55.3& 87.8 &71.1& 61.1& 46.5& 22.1 &6.6 &\textbf{55.3}& 48.9\\
			GMNet \citep{zhou2021gmnet}&57.3  &86.5& 73.1& 61.7& 44.0&\textbf{ 42.3} &14.5& 48.7 &47.4\\
			EGFNet \citep{10234530}&57.5& 89.8&71.6&63.9&46.7&31.3&6.7&52.0&57.4\\
			DPLNet \citep{dong2024efficient}&59.3&-&-&-&-&-&-&-&-\\
			CMX \citep{zhang2023cmx}&59.7& 90.1& 75.2& 64.5& 50.2& 35.3 &8.5& 54.2& 60.6\\
			CRM-RGBT-Seg \citep{shin2024complementary}&61.4& 90.0& 75.1& 67.0& 45.2& 49.7& \textbf{18.4}& 54.2& 54.4\\
			\midrule
			MFNet(baseline) \citep{ha2017mfnet}&63.5 &92.6&82.1&78.2&89.6&24.1&1.2&46.2&94.2\\
			MS2Fusion-MFNet&\textbf{66.3}&\textbf{94.4}&\textbf{82.5}&\textbf{81.0}&\textbf{89.9}&34.7&0.0&	49.7&	\textbf{98.3}\\
			\bottomrule
	\end{tabular}}
	\label{MFNet_com}
\end{table*}

\textbf{Comparison on the SemanticRT dataset.} As shown in Table \ref{SemanticRT_com}, our MS2Fusion method also exhibits superior performance. Specifically, it achieves an improvement of 1.0\% over the baseline and differs by only 0.5\% from the current state-of-the-art model. These findings highlight the robustness and versatility of our approach in multispectral feature fusion, proving its adaptability to various tasks.

Based on the experimental comparisons on the two RGB-T semantic segmentation benchmarks, we demonstrates that our MS2Fusion is also equally effective, with a high degree of generality and adaptability to a wide range of downstream tasks.\par

\begin{table*}[htp]
	\centering
	\footnotesize
	\caption{Comparison on the SemanticRT dataset.}
	\resizebox{\linewidth}{!}{
		\begin{tabular}{rcccccccccccccc}
			\toprule
			Methods&mIoU&CarStop&Bike&Bicyclist&Mtcycle&Mtcyclist&Car&Tricycle&TrafLight&Box&Pole&Curve&Person\\
			\midrule
			PSTNet \citep{shivakumar2020pst900}&68.0&71.1&62.3&58.5&47.3&55.2&85.4&44.2&75.7&83.0&71.7&62.2&72.2\\
			RTFNet \citep{sun2019rtfnet}&75.5 &79.6&68.0&67.4&63.7&61.6&90.4&66.0&78.3&85.9&78.0&67.2&78.9\\
			EGFNet \citep{10234530}&77.4&78.6&71.3&70.9&68.4&66.1&90.5&71.5&80.4&85.4&76.5&66.9&83.7\\
			ECM \citep{ji2023semanticrt}&\textbf{79.3}&\textbf{80.2}&75.0&\textbf{75.5}&71.4&\textbf{70.4}&90.3&\textbf{74.0}&85.9&85.6&77.2&68.3&85.0\\
			\midrule
			MFNet(baseline) \citep{ha2017mfnet}&77.8&75.3&77.1&63.2&71.1&57.3&\textbf{97.9}&66&85.9&89.5&\textbf{82.4}&\textbf{85.0}&83.2\\
			MS2Fusion-MFNet&78.8&75.3&\textbf{77.7}&66.3&\textbf{72.3}&59.4&\textbf{97.9}&68.2&\textbf{86.0}&\textbf{89.6}&81.8&84.6&\textbf{86.7}\\
			\bottomrule
	\end{tabular}}
	\label{SemanticRT_com}
\end{table*}

\begin{table*}[!t]
	\centering
	\footnotesize
	\caption{Comparison on the VT821, VT1000 and VT5000 datasets. 
		($\downarrow$ indicates smaller value is better, while $\uparrow$ indicates larger value is better.)}
        \resizebox{\linewidth}{!}{
		\begin{tabular}{r|c|c|c|c|c|c|c|c|c|c|c|c}
			\toprule
			\multirow{2}{*}{Methods} &\multicolumn{4}{c|}{VT821} &\multicolumn{4}{c|}{VT1000} &\multicolumn{4}{c}{VT5000}\\ \cmidrule{2-13}
			&S$\uparrow$&adpE$\uparrow$&adpF$\uparrow$&MAE$\downarrow$&S$\uparrow$&adpE$\uparrow$&adpF$\uparrow$&MAE$\downarrow$&S$\uparrow$&adpE$\uparrow$&adpF$\uparrow$&MAE$\downarrow$\\ 
			\midrule
			MTMR \citep{wang2018rgb}&72.5&81.5&66.2&10.9&70.6&83.6&71.5&11.9&68.0&79.5&59.5&11.4\\
			M3S-NIR \citep{2019M3S}&72.3&85.9&76.4&14.0&72.6&82.7&71.7&14.5&65.2&78.0&57.5&16.8\\
			SGDL \citep{8744296}&76.5&84.7&76.1&8.5&78.7&85.6&76.4&9.0&75.0&82.4&67.2&8.9\\
			PoolNet \citep{2019A}&75.1&73.9&57.8&10.9&83.4&81.3&71.4&6.7&76.9&75.5&58.8&8.9\\
			R$^3$Net\citep{ijcai2018p95}&78.6&80.9&66.0&7.3&84.2&85.9&76.1&5.5&75.7&79.0&61.5&8.3\\
			CPD \citep{8953521}&82.7&83.7&71.0&5.7&90.6&90.2&83.4&3.2&84.8&86.7&74.1&5.0\\
			MMCI \citep{CHEN2019376}&76.3&78.4&61.8&8.7&88.6&89.2&80.3&3.9&82.7&85.9&71.4&5.5\\
			AFNet \citep{2019Adaptive}&77.8&81.6&66.1&6.9&88.8&91.2&83.8&3.3&83.4&87.7&75.0&5.0\\
			TANet \citep{2019Three}&81.8&85.2&71.7&5.2&90.2&91.2&83.8&3.0&84.7&88.3&75.4&4.7\\
			S2MA \citep{2020Learning}&81.1&81.3&70.9&9.8&91.8&91.2&84.8&2.9&85.3&86.4&74.3&5.3\\
			JLDCF \citep{2021Siamese}&83.9&83.0&72.6&7.6&91.2&89.9&82.9&3.0&86.1&86.0&73.9&5.0\\
			FMCF \citep{2020RGB}&76.0&79.6&64.0&8.0&87.3&89.9&82.3&3.7&81.4&86.4&73.4&5.5\\
			ADF \citep{9767629}&81.0&84.2&71.7&7.7&91.0&92.1&84.7&3.4&86.4&89.1&77.8&4.8\\
			MIDD \citep{2021Multi}&87.1&89.5&80.3&4.5&91.5&93.3&88.0&2.7&86.8&89.6&79.9&4.3\\
			LSNet \citep{10042233}&87.2&91.0&81.4&3.6&92.1&94.9&88.1&2.3&87.5&92.0&81.7&3.7\\
			CAVER \citep{10015667, PENG2024410} &89.8&92.8&\textbf{87.7}&\textbf{2.7}&93.6&94.9&91.1&1.7&89.9&94.1&84.9&\textbf{2.8}\\
			DPLNet \citep{dong2024efficient}&87.8&90.8&81.0&4.3&92.8&95.1&88.1&2.2&87.9&91.6&82.8&3.8\\
            \midrule
			MSEDNET(baseline) \citep{PENG2024410}&87.6&89.7&80.3&3.9&92.8&93.9&86.8&2.2&88.1&91.6&82.1&3.7\\
			MS2Fusion-MSEDNET &\textbf{90.4}&\textbf{93.5}&86.3&3.2&\textbf{94.4}&\textbf{97.2}&\textbf{91.8}&\textbf{1.6}&\textbf{90.2}&\textbf{94.2}&\textbf{86.4}&3.0\\
			
			\bottomrule
	\end{tabular}}
	\label{VT_comparision}
\end{table*}

\subsubsection{Experiments on RGB-T Salient Object Detection (RGB-T SOD)} \label{subsubsec4.4.2}
\textbf{Evaluation metrics:}
Following \citep{borji2019salient}, Mean Absolute Error (MAE), F-Measure (adpF), S-Measure (S), and E-Measure (adpE) are commonly used as evaluation metrics for SOD.\par

\textbf{VT821 dataset \citep{wang2018rgb}:}  It provides 821 aligned RGB-T image pairs capturing challenging real-world scenarios. It specifically includes illumination variations, object occlusions, and low-contrast thermal conditions to test model robustness in dynamic settings.

\textbf{VT1000 dataset \citep{8744296}:} It comprises 1,000 pairs of RGB-T images, offering a broader range of scenes, including urban, rural, indoor, and outdoor environments. This dataset enhances diversity by covering different weather conditions, times of day (daylight and nighttime), and various object types (e.g., pedestrians, vehicles, animals). 

\textbf{VT5000 dataset \citep{9767629}:} It is a large-scale dataset with 5,000 pairs of RGB-T images. 
It features a wide array of challenging conditions, such as extreme weather, dense occlusion, and multi-object scenes, offering high diversity and difficulty. 

MSEDNET \citep{PENG2024410} is employed as our baseline RGB-T SOD method due to its proven hierarchical fusion architecture and superior performance, and MS2Fusion-MSEDNET denotes the baseline equipped with our proposed MS2Fusion module. 
As shown in \Cref{VT_comparision}, the MS2Fusion-MSEDNET method achieves state-of-the-art performance for RGB-T SOD on the VT821, VT1000 and VT5000 datasets as well. It is clear to observe that MS2Fusion-MSEDNET ranks first or second on all evaluation metrics, demonstrating its superior capabilities of feature fusion in the RGB-T SOD task.

\subsection{Ablation Studies}
\label{subsec4.5}
All ablation studies in this paper are conducted on the FLIR dataset. Unless otherwise specified, the experimental setting, model architecture, and parameter settings follow those described in \Cref{subsec4.2}. 
\subsubsection{Different Detection Frameworks}
\label{subsubsec4.5.1}
To thoroughly evaluate the generalization of the MS2Fusion module, comparative experiments are conducted on two mainstream detection frameworks, YOLOv5 and CoDetr \citep{zhou2024optimizing}, with add fusion serving as the baseline method. \Cref{difframe} unequivocally demonstrate that our fusion approach achieves remarkable performance improvements in both frameworks, delivering a 3.3\% boost in mAP@0.5 over the baseline. These findings strongly validate the efficacy and versatility of MS2Fusion. Notably, the module exhibits excellent plug-and-play characteristics, enabling seamless integration with various detection frameworks and providing a flexible yet powerful solution for multimodal object detection tasks.

\begin{table*}[htp]
	\centering
	\footnotesize
	\caption{Effect of different backbones and detection frameworks.}
	\resizebox{!}{!}{
		\begin{tabular}{cccllllll}
			\toprule
			Num & Framework&Fusion Model & Person & Car & Bicycle & mAP@0.5 & FPS\\
			\midrule
			1&\multirow{2}{*}{YOLOv5}&Baseline &83.8&	88.9&	67.3&80.0&43.2 \\
			2 & &MS2Fusion&85.1&	89.8&	74.9&	83.3 (+3.3) &24.4\\
			\midrule 
			3& \multirow{2}{*}{CoDetr}&Baseline &88.0	&91.7	&73.8	&84.5&5.9\\ 
			4 & &MS2Fusion& 90.2	&93.4	&79.6	&87.8 (+3.3)&4.8\\
			\bottomrule
		\end{tabular}
		
		}
	\label{difframe}
\end{table*}

\subsubsection{Different Backbones}
\label{subsubsec4.5.2}
To comprehensively evaluate the architectural compatibility of the MS2Fusion module, we conducted systematic experiments with three distinct backbone networks: VGG16 \citep{simonyan2014very}, ResNet50 \citep{he2016deep}, and CSPDarkNet53 \citep{redmon2016you}. As demonstrated in Table \ref{difbackbone}, MS2Fusion achieves consistent performance gains across all architectural configurations. Specifically, our method yields improvements of 1.0\%, 4.8\%, and 3.3\% over the baseline when employing VGG16, ResNet50, and CSPDarkNet53 backbones, respectively. These experimental results not only validate the effectiveness of the MS2Fusion module but also substantiate its remarkable generalization capability across diverse backbone architectures.

\begin{table*}[htp]
	\centering
	\footnotesize
	\caption{Effect of different backbones.}
	\resizebox{!}{!}{
		\begin{tabular}{ccclllll}
			\toprule
			Num & Backbone&Fusion Model & Person & Car & Bicycle & mAP@0.5 \\
			\midrule
			1& \multirow{2}{*}{VGG16}&Baseline &78.9&	87.7&	51.2&	72.6\\
			2& & MS2Fusion&79.0&	87.8&	53.8&	73.6 (+1.0)\\
			\midrule 
			3& \multirow{2}{*}{ResNet50}&Baseline &76.8	&85.7	&44.6	&69.0\\ 
			4 & &MS2Fusion& 80.2	&88.4	&52.8	&73.8 (+4.8)\\
			\midrule
			5&\multirow{2}{*}{CSPDarkNet53}&Baseline &83.8&	88.9&	67.3&80.0 \\
			6 & &MS2Fusion&85.1&	89.8&	74.9&	83.3 (+3.3)\\
			\bottomrule
		\end{tabular}
		
		}
	\label{difbackbone}
\end{table*}

\begin{table*}[htp]
	\centering
	\footnotesize
	\caption{Effect of different fusion modules.}
	\resizebox{\linewidth}{!}{
		\begin{tabular}{cccclllll}
			
			\toprule
			Num & CP-SSM & SP-SSM & FF-SSM& Person & Car & Bicycle & mAP@0.5 &Params(M)  \\
			\midrule
			1	&  	&   &  &83.8 &88.9 &67.3 &80.0 &72.7 \\
			2	&\Checkmark	&   & &83.8 &89.8 &72.7 &82.1 (+2.1) &84.5 \\
            3	&	&\Checkmark   & &85.6&90.4&70.6&82.2 (+2.2)&90.4  \\
			4	&	&   &\Checkmark &84.9& 	90.1& 	72.4& 	82.5 (+2.5)&86.0  \\
			
            5   &\Checkmark	&\Checkmark   &  &84.9 	&90.3	&72.9	&82.7 (+2.7) &117.0\\ 
            6	&\Checkmark	&   &\Checkmark  &85.5 	&90.0	&72.5	&82.7 (+2.7) &97.8\\
            7	&&\Checkmark	  &\Checkmark  &85.5 	&90.3	&72.9	&82.9 (+2.9) &103.7\\
			8	&\Checkmark	& \Checkmark  &\Checkmark &85.1 &89.8 &74.9 &83.3 (+3.3) &130.2\\
			\bottomrule
		\end{tabular}
		}
	\label{difmodules}
\end{table*}

\subsubsection{Different Modules}
\label{subsubsec4.5.3}
Our comparative study of CP-SSM, SP-SSM, and FF-SSM modules reveals their complementary strengths in object detection. CP-SSM excels in enhancing modality-specific features, improving mAP by 2.1\% and boosting challenging category detection (e.g., bicycles) by 5.4\%. SP-SSM focuses on extracting cross-modal shared features with strong generalization, while FF-SSM delivers efficient heterogeneous feature fusion, yielding a 2.5\% mAP gain with minimal overhead and consistent performance across categories.

Combining modules further amplifies performance. SP-SSM + FF-SSM achieves 82.9\% mAP, while integrating all three modules pushes mAP to 83.3\%, improving bicycle detection by 7.6\%. This synergy balances feature complementarity, sharing, and fusion, enhancing accuracy while keeping parameter growth reasonable.

In summary, CP-SSM, SP-SSM, and FF-SSM work in concert to boost detection via modality-specific enhancement, cross-modal generalization, and adaptive fusion—preserving unique features, uncovering shared semantics, and reducing redundancy effectively.

\subsubsection{Different Fusion Layers}
\label{subsub4.5.4}
As shown in Table \ref{difstages}, the incremental addition of MS2Fusion modules across different feature stages yields distinct performance characteristics. Initial test at only the P3 stage (Row 2) demonstrates a trade-off effect, where 'bicycle' AP experiences a substantial 5.7\% improvement to 71.4\% at the cost of marginal decreases in 'person' (83.4\%) and 'car' (88.5\%) detection, ultimately elevating the mAP to 81.1\%. 
Expanding the fusion to both P3 and P4 stages (Row 3) reverses this pattern, boosting 'person' and 'car' AP to 85.8\% and 90.1\% 
When employing fusion at all three stages (Row 4), it achieves optimal balance. While maintaining strong performance on 'person' (85.1\%) and 'car' (89.8\%), it delivers a remarkable 74.9\% AP for 'bicycles', the highest among all configurations and pushes the overall mAP to 83.3\%. 
This systematic evaluation clearly demonstrates that multi-level feature fusion enables more effective cross-modal feature integration, with full-stage implementation proving particularly advantageous for challenging categories like 'bicycles' while preserving performance on other objects.

\begin{table*}[!t]
	\centering
	\footnotesize
	\caption{Effect of fusion at different layers.}
        \resizebox{!}{!}{
	\begin{tabular}{cccccccl}
		\toprule
		Number & P3 & P4 & P5& Person & Car & Bicycle & mAP@0.5\\
		\midrule
		1	& & & & 83.8 &88.9 &67.3 &80.0 \\
		2	&\Checkmark	& &&83.4&	88.5&71.4&	81.1 (+1.1)\\
		3	&\Checkmark	& \Checkmark & & 85.8	&90.1	&70.0&82.0 (+2.0) \\
		4	&\Checkmark	& \Checkmark  &\Checkmark &85.1&89.8&74.9&83.3 (+3.3)\\
		\bottomrule
	\end{tabular}}
	\label{difstages}
\end{table*}

\begin{table*}[!t]
	\centering
	\footnotesize
	\caption{Finetunes in CP-SSM.}
        \resizebox{!}{!}{
	\begin{tabular}{clcccc}
		\toprule
		Num & Finetune &  Person & Car & Bicycle & mAP@0.5\\
		\midrule
		1	&rows &85.1&89.8&74.9&83.3 \\
		2	&columns &64.1&	90.3	&72.8	&82.4  \\
		3	&rows and columns &85.4&	90.4&	70.2&	82.0\\
		\midrule
		4	&w/o exchange $C$ &85.7&	90.6&71.7&82.7\\
		5	&exchange $C$ &85.1&89.8&74.9&83.3 \\
		\bottomrule
	\end{tabular}}
	\label{finetune}
\end{table*}

\subsubsection{Discussion on CP-SSM}
\label{subsubsec4.5.5}
\textbf{Different expanding methods.} Following VMamba \citep{liu2024vmamba}, we also explore various scanning orientations for SSM in Table \ref{finetune}. 
As shown in Table \ref{finetune}(row $1\sim3$), we find that multidirectional scanning adversely impacts experimental outcomes, rendering it unsuitable for object detection tasks. We think that multidirectional scanning may alter object features, contradicting the stability required for accurate object detection. \par

\textbf{Exchange \texorpdfstring{$C$}{}-parameters.}
As shown in Table \ref{finetune}, the experimental study of the cross-branch C-parameter exchange mechanism in the CP-SSM module demonstrates that exchanging hidden state projection matrices can significantly improve model performance. Compared to non-exchange configurations, this mechanism achieves a stable performance improvement of 0.6\%, not only enhancing the feature capture capability for complex data relationships but also optimizing cross-feature interaction efficiency. Experimental validation shows that this parameter exchange strategy is particularly effective in scenarios requiring fine-grained data representation, enabling the CP-SSM module to demonstrate outstanding performance in cross-modal feature extraction tasks. This finding highlights the importance of parameter interaction mechanism design in modern feature learning architectures.

\subsubsection{Comparison with Different Input Modalities}
\label{subsubsec4.5.6}
\begin{table}[!t]
	\centering
	\footnotesize
	\caption{Performance of different inputs. (V, T denotes Visible and thermal, respectively. V+T represents the input with dual modalities, while V+V or T+T denotes input with a single modality.)}
	\begin{tabular}{cccl}
		\toprule
		Num & Model &Input & mAP@0.5 \\
		\midrule
		1& \multirow{2}{*}{YOLOv5}&V &67.8\\
		2& &T&73.9\\
		\midrule
		3& \multirow{3}{*}{Baseline}&V+V &61.2\\
		4 & &T+T& 77.8\\
		5 & &V+T& 80.0\\
		\midrule
		6& \multirow{3}{*}{Ours}&V+V &68.4 (+7.2)\\
		7 & &T+T&82.1 (+4.3) \\
		8 & &V+T&83.3 (+3.3) \\
		\bottomrule
	\end{tabular}
	\label{difinput}
\end{table}
Table \ref{difinput} evaluates the performance of our model under conditions when certain input modalities are missing.
The results demonstrate that our model can achieve competitive results even when only one modality (V+V or T+T) is used. Specifically, when only thermal images are provided, our model's performance is just 1.2\% lower than that achieved with multi-modal inputs (as seen in Rows 7 and 8). 
This highlights the robustness of our approach in handling incomplete modality inputs. Additionally, our method shows substantial improvements over the baseline when the same two modalities are used (comparing Rows 3 and 6, 4 and 7, 5 and 8). 
This improvement underscores the effectiveness of our module in exploiting the shared spatial features of heterogeneous data. 
By enhancing the generalization of cross-modal features, our model effectively utilizes information from one available modality to enhance another modality's features. 
These findings indicate that our approach not only maintains high performance with reduced input data but also significantly leverages the shared and complementary features across different modalities. 

\begin{table}[t]
    \centering
    \footnotesize
    \caption{Input Configuration for FF-SSM Module}
    \begin{tabular}{ccccc}
    \toprule
        FF-SSM input & mAP@0.5 & person & car & bicycle  \\ 
        \midrule
        (V, T) & 83.1 & 85.4 & 90.2 & 73.9  \\
        (T, V) & 82.5 & 85.1 & 90.1 & 72.4  \\
        ((V, T), (T, V)) & 83.3 & 85.1 & 89.8 & 74.9 \\
        \bottomrule
        \label{FF-SSM_input}
    \end{tabular}
\end{table}

\begin{figure*}[t]
	\centering
	\subfloat[(V, T))]{\includegraphics[width=0.3\linewidth]{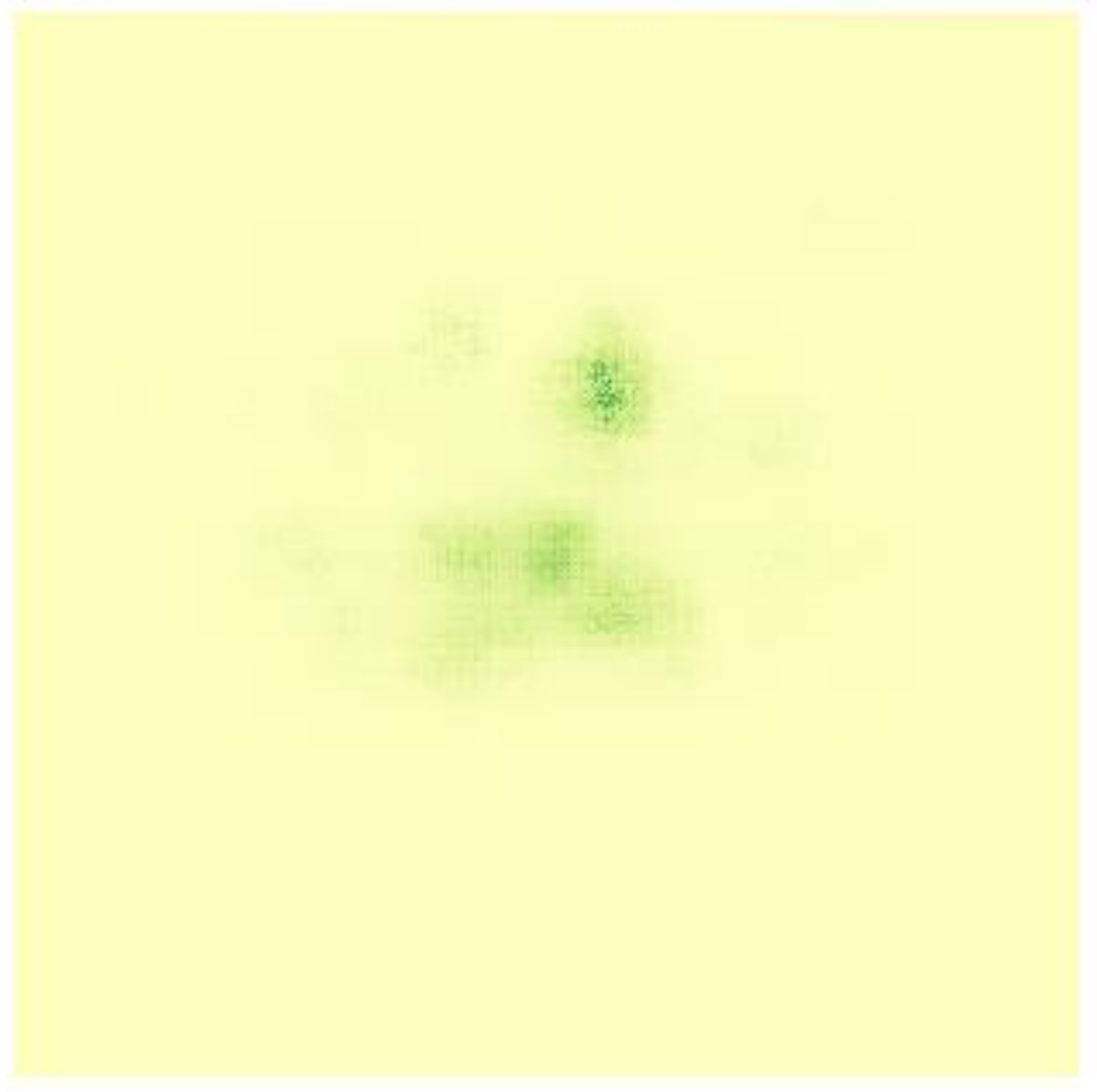}%
		\label{vt_erf}}
	\hfil
	\subfloat[(T, V)]{\includegraphics[width=0.3\linewidth]{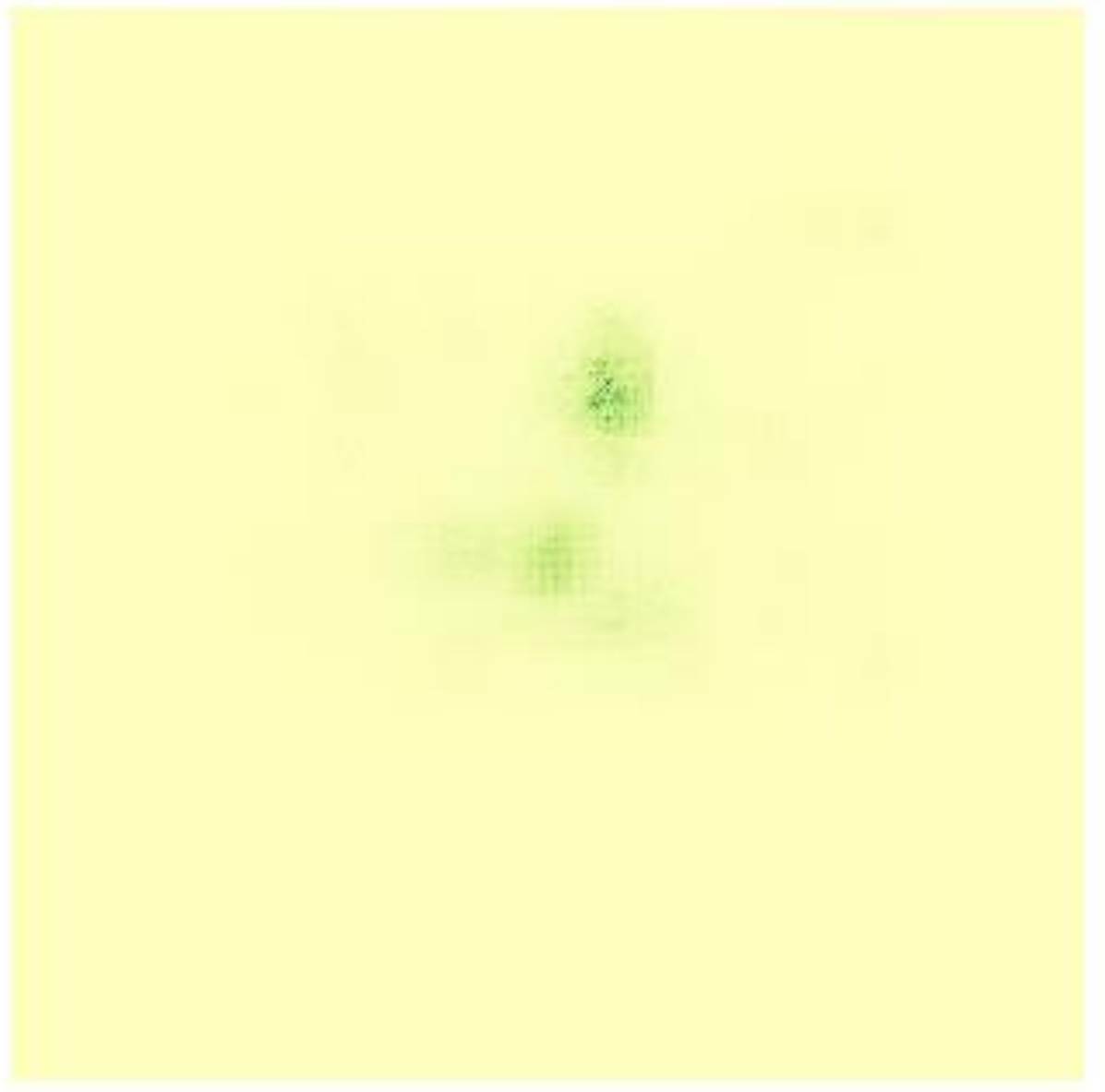}%
		\label{tv_erf}}
        \hfil
        \subfloat[((V, T), (T, V))]{\includegraphics[width=0.3\linewidth]{MS2Fusion_erf.pdf}%
		\label{vttv_erf}}
	\caption{ ERF visualizations comparing different input configurations of the FF-SSM module: (a) unidirectional (V, T) input, (b) unidirectional (T, V) input, and (c) our proposed bidirectional ((V, T), (T, V)) input. The ERF map demonstrates that the bidirectional strategy achieves a significantly broader receptive field compared to the unidirectional approaches. }
	\label{ff_input_erf}
\end{figure*}

\subsubsection{Analysis of FF-SSM Input Configurations}
\label{subsubsec4.5.7}
In this section, we systematically evaluate the FF-SSM module with three distinct input configurations: (V, T), (T, V), and bidirectional ((V, T), (T, V)). As shown in \Cref{FF-SSM_input}, single-order inputs exhibit significant performance variations: the (V, T) configuration achieves 83.1\% mAP@0.5 (0.2\% decrease from baseline), with particularly pronounced accuracy degradation for 'bicycle' detection (1.0\% drop). The (T, V) configuration shows more substantial performance decline, reaching only 82.5\% mAP@0.5 (0.8\% decrease) and suffering a 2.5\% accuracy reduction for 'bicycle' detection.
These results reveal a critical feature attenuation phenomenon, where early input features are gradually forgotten during state propagation. To overcome this limitation, we propose a novel bidirectional architecture that establishes robust cross-modal feature memory pathways. Our solution not only elevates overall performance to 83.3\% mAP@0.5 but also delivers significant improvements in 'bicycle' detection accuracy (74.9\%), while maintaining stable detection precision for both 'person' and 'car' categories.

As illustrated in \Cref{ff_input_erf}, we conducted a comprehensive comparison of three ERF map configurations for the FF-SSM module: (V, T), (T, V), and bidirectional ((V, T), (T, V)). The results demonstrate that our bidirectional input strategy yields a substantially expanded effective receptive field compared to unidirectional approaches.
This empirical evidence strongly corroborates the efficacy of our proposed bidirectional architecture in enhancing the model's receptive field capacity. The innovative design successfully mitigates feature attenuation in state-space models while maintaining linear computational complexity, thereby providing an efficient and effective solution for cross-modal feature fusion.

\subsection{Qualitative Analysis}
\label{subsec4.6}
\subsubsection{Statistical Analysis of the Model}
\label{subsubsec4.6.1}
\begin{longtable}[c]{@{}ccccc@{}}
\caption{Statistical analysis of the model( Perform the experiment 10 times with random seeds released.).}
\label{statiscal_analyse}\\
\toprule
\multirow{2}{*}{} & \multicolumn{2}{c}{Mean(\%)} & \multicolumn{2}{c}{VAR(\%)} \\
                  & mAP@0.5        & mAP         & mAP@0.5        & mAP        \\ \midrule
\endfirsthead
\endhead
\bottomrule
\endfoot
\endlastfoot
FLIR              & 82.99          & 40.60       & 0.15           & 2.47       \\
LLVIP             & 97.41          & 63.46       & 0.03           & 2.31       \\
M$^3$FD              & 89.11          & 59.76       & 0.18           & 0.42       \\ \bottomrule
\end{longtable}
The model demonstrates excellent robustness and stability based on the analysis of 10 random seed experiments as shown in \Cref{statiscal_analyse}. On the FLIR dataset, it achieves an mAP@0.5 of 82.99\% with a variance of only 0.15\%. Its performance on the LLVIP dataset is particularly outstanding, reaching an mAP@0.5 of 97.41\% with an extremely low variance of 0.03\%. On the M$^3$FD dataset, the model attains an mAP@0.5 of 89.11\% and an mAP variance of just 0.42\%. These results indicate that the model exhibits remarkable stability, fully demonstrating its strong resistance to variations in initial conditions. It maintains consistent performance across different data distributions, showcasing excellent practical engineering value.
\subsubsection{Heatmap Visualization}
\label{subsubsec4.6.2}
\begin{figure*}[!t]
	\centering
	\includegraphics[width=1\linewidth]{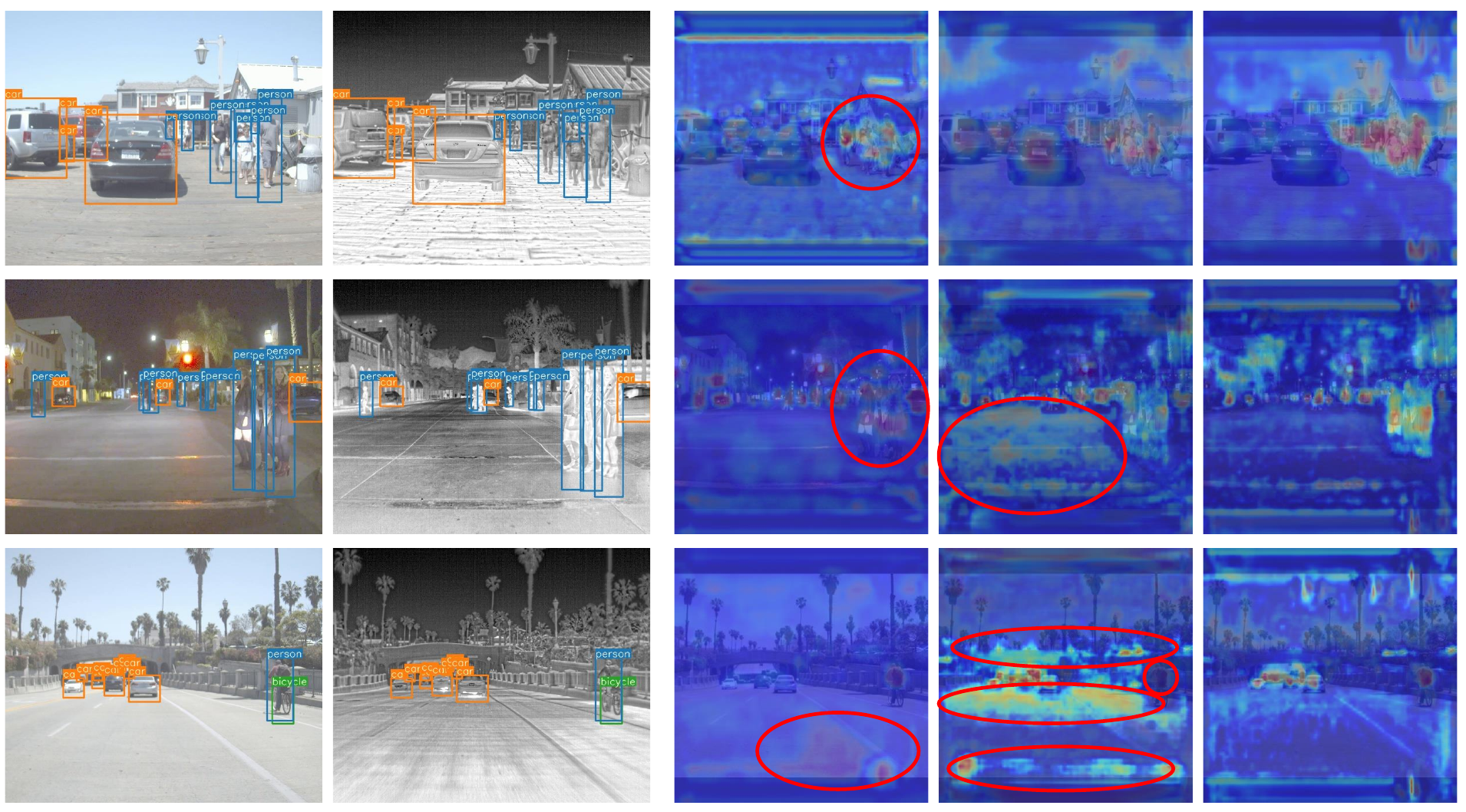}
	\caption{Heatmap visualization. (The first and second columns are visible and thermal images; The third, fourth and fifth columns are heatmaps of baseline, ICAFusion and MS2Fusion, respectively.)}
	\label{heatmap}
\end{figure*}

\Cref{heatmap} presents the heatmap comparisons of three competing methods (baseline, ICAFusion and MS2Fusion) on RGB-T images. Through comprehensive analysis across multiple scenarios, the MS2Fusion method demonstrates consistent superiority.
In the parking lot scenario (first row), the baseline method manages to detect partial instances of cars and pedestrians, yet produces significantly incomplete bounding boxes. Although ICAFusion shows noticeable improvement by capturing more objects, it still suffers from occasional missed detections. By contrast, MS2Fusion achieves near-perfect performance, precisely localizing all cars and pedestrians with highly accurate bounding boxes.
The performance gap becomes even more pronounced in the challenging nighttime street scene (second row). While the baseline method can identify some pedestrians and vehicles, its detection accuracy proves inadequate. ICAFusion offers moderate improvement over the baseline, yet still struggles with low detection accuracy in these low-light conditions. Remarkably, MS2Fusion maintains excellent performance, reliably identifying all objects with exceptional precision even in this demanding scenario.
Similarly, in the road scene (third row), the baseline method exhibits several limitations, including missed detections of vehicles and cyclists. ICAFusion partially addresses these issues by detecting more instances, but its accuracy remains suboptimal. Impressively, MS2Fusion again outperforms both competitors, achieving better detection performance with precise bounding boxes and minimal false negatives. \par

The MS2Fusion method demonstrates superior performance across all evaluated scenarios, achieving both high detection accuracy and precise object localization. Quantitative and qualitative analyses reveal three key advantages: (1) Compared to CNN-based baseline methods, MS2Fusion captures broader object regions through its expanded receptive field; (2) Relative to Transformer-based ICAFusion, it achieves more precise contour fitting; and (3) It maintains the lowest false negative rate, particularly in challenging low-light conditions. These improvements stem from MS2Fusion's effective cross-modal fusion mechanism, which optimally combines complementary information from RGB and thermal modalities. The heatmap visualizations further confirm that MS2Fusion's hybrid architecture successfully balances wide coverage (CNN advantage) and precise localization (Transformer strength), establishing it as an effective solution for multispectral object detection.

\subsubsection{Visualization of Feature Fusion Comparisons}
\label{subsubsec4.6.3}
\begin{figure*}[!htb]
	\centering
	\includegraphics[width=0.8\linewidth]{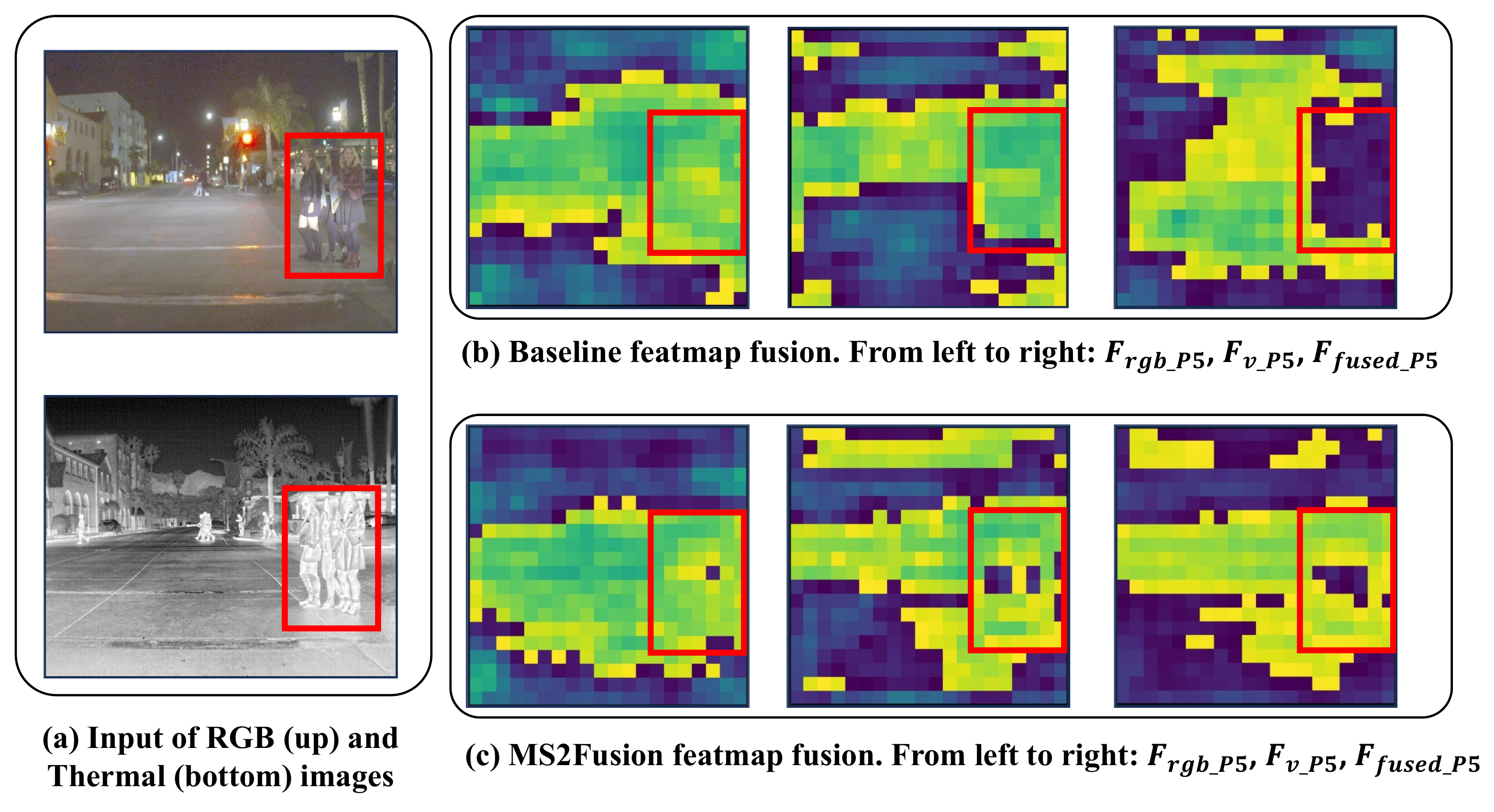}
	\caption{Visualization of different feature map fusion.}
	\label{featmap visualization}
\end{figure*}

We selected the P5 layer features for visualization, where $F_{rgb\_p5}, F_{v\_p5}$, $F_{fused\_p5}$ denote the RGB features, Thermal features, and fused features, respectively. \Cref{featmap visualization} demonstrates the superior performance of MS2Fusion compared to the baseline approach, particularly in feature preservation and enhancement. 
The baseline method suffers from critical limitations: (1) it simply superimposes RGB and thermal feature maps through direct summation, often causing information loss or modal conflicts; (2) it fails to effectively leverage multimodal complementarity. These shortcomings are evident in its suboptimal feature representations.

In contrast, MS2Fusion addresses these issues through a sophisticated multi-stage fusion framework comprising three key modules. This hierarchical architecture enables selective retention of the most discriminative features from each modality while suppressing redundancy, as clearly visualized in the red rectangle of \Cref{featmap visualization}. 

The framework's advantages manifest in enhanced feature retention that preserves modality-specific details, optimal complementarity utilization that dynamically balances RGB and thermal contributions, and improved scene adaptability that maintains robustness in complex environments. 
Quantitative results also confirm that MS2Fusion generates richer, clearer feature maps that directly translate to superior performance, resolving the fundamental trade-off between feature preservation and cross-modal integration that plagues the other methods.

\subsubsection{Visual Analysis of Shared and Complementary Features.}
\label{subsubsec4.6.4}
\begin{figure*}[!htb]
	\centering
	\includegraphics[width=.8 \linewidth]{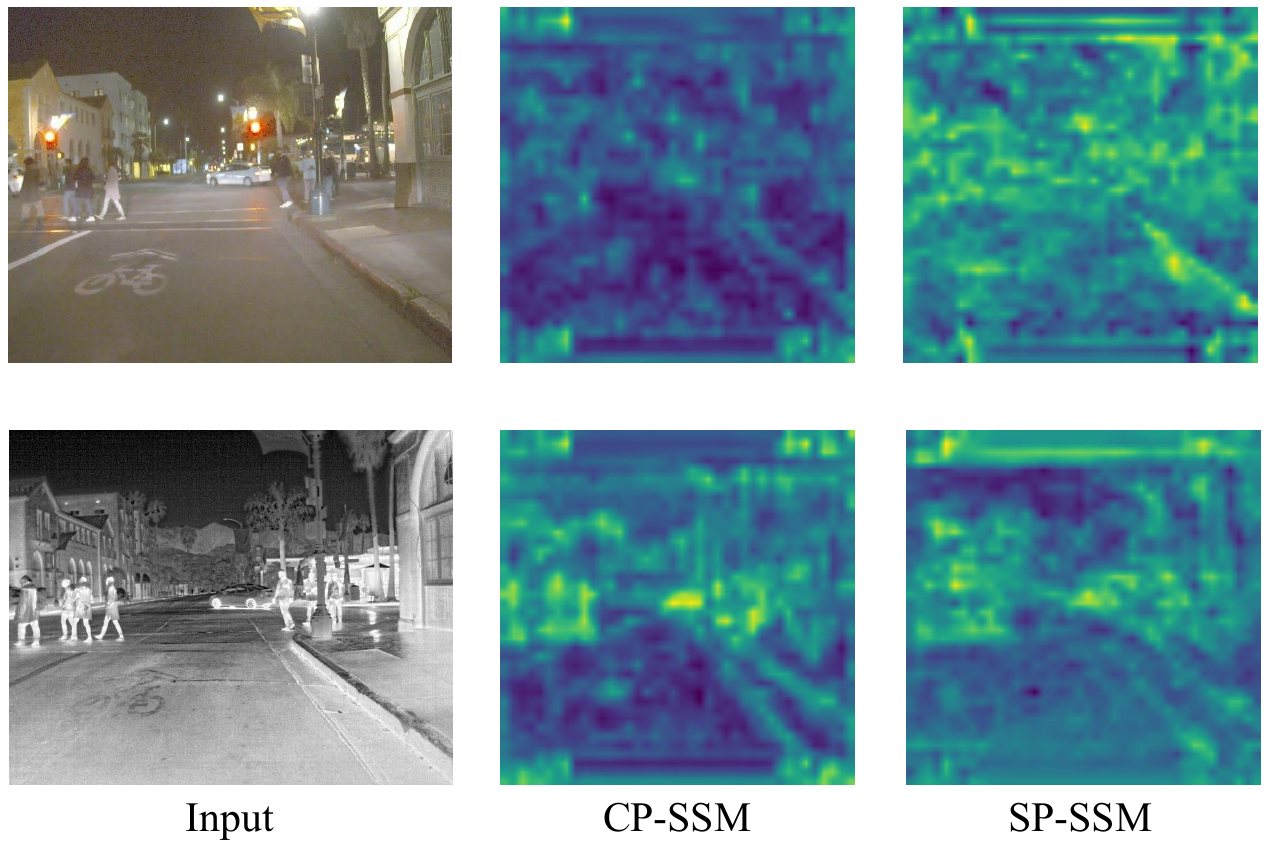}
	\caption{Visualization of intermediate features under low-light conditions. The first column shows RGB and thermal inputs; the second and third columns display complementary (CP-SSM) and shared (SP-SSM) features, respectively.}
	\label{feature_vis}
\end{figure*}

As illustrated in \Cref{feature_vis}, the first column shows the RGB and thermal input images, while the second and third columns visualize the complementary and shared feature maps learned by CP-SSM and SP-SSM, respectively.
Due to the low-light conditions of this scene, the RGB modality captures relatively weak and sparse features, whereas the thermal modality provides richer structural and intensity cues. The CP-SSM focuses on these modality-specific differences, emphasizing distinctive thermal gradients and texture variations.
In contrast, the SP-SSM effectively fuses information from both modalities, producing shared semantic representations that consistently highlight object contours, pedestrian regions, and road boundaries across RGB and thermal domains. This indicates that SP-SSM compensates for illumination-induced degradation in the visible spectrum by enforcing semantic consistency and cross-modal alignment.
Overall, this visualization demonstrates that MS2Fusion’s dual-path design enables robust feature learning under challenging night-time conditions, balancing modality complementarity and shared semantic understanding.

\subsubsection{Visualization of RGB-T Semantic Segmentation Samples}
\label{subsubsec4.6.5}
\begin{figure*}[!htb]
	\centering
	\includegraphics[width=1 \linewidth]{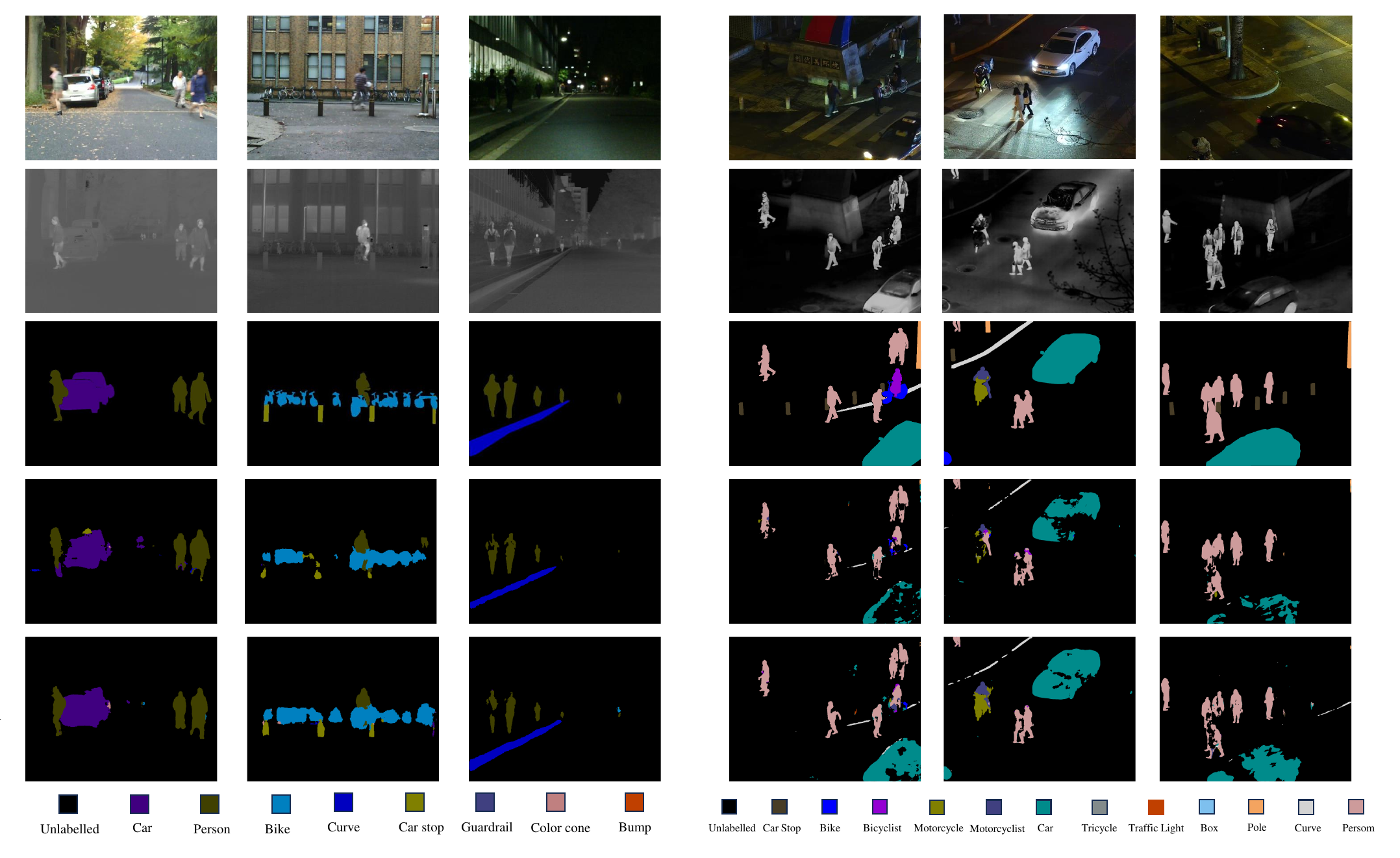}
	\caption{Visualization of semantic segmentation of our model on MFNet and SemanticRT datasets. (The visible, thermal and groundtruth are provided in the first three rows, while the result of the baseline and MS2Fusion methods are shown in the fourth and fifth rows.)}
	\label{semantic_seg}
\end{figure*}

\Cref{semantic_seg} presents comparative semantic segmentation results across different input modalities and models. 
The visualization consists of five distinct rows: (1) visible input, (2) thermal input, (3) groundtruth annotations serving as reference labels, (4) baseline model predictions, and (5) MS2Fusion results. 
The baseline model exhibits noticeable deficiencies, particularly in handling pedestrian and vehicle boundaries, where segmentation appears blurred and misclassified. 
However, MS2Fusion demonstrates marked improvement in these challenging areas. The proposed method shows particular strength in preserving fine details along object contours and maintaining segmentation consistency in complex backgrounds, as evidenced by its precise delineation of persons and vehicles. 
Quantitative analysis confirms that MS2Fusion's multimodal fusion strategy effectively enhances segmentation accuracy by reducing classification errors and improving overall prediction quality compared to the baseline approach.

\subsubsection{Visualization of RGB-T Salient Object Detection Samples}
\label{subsubsec4.6.6}
\begin{figure*}[!htb]
	\centering
	\includegraphics[width=1 \linewidth]{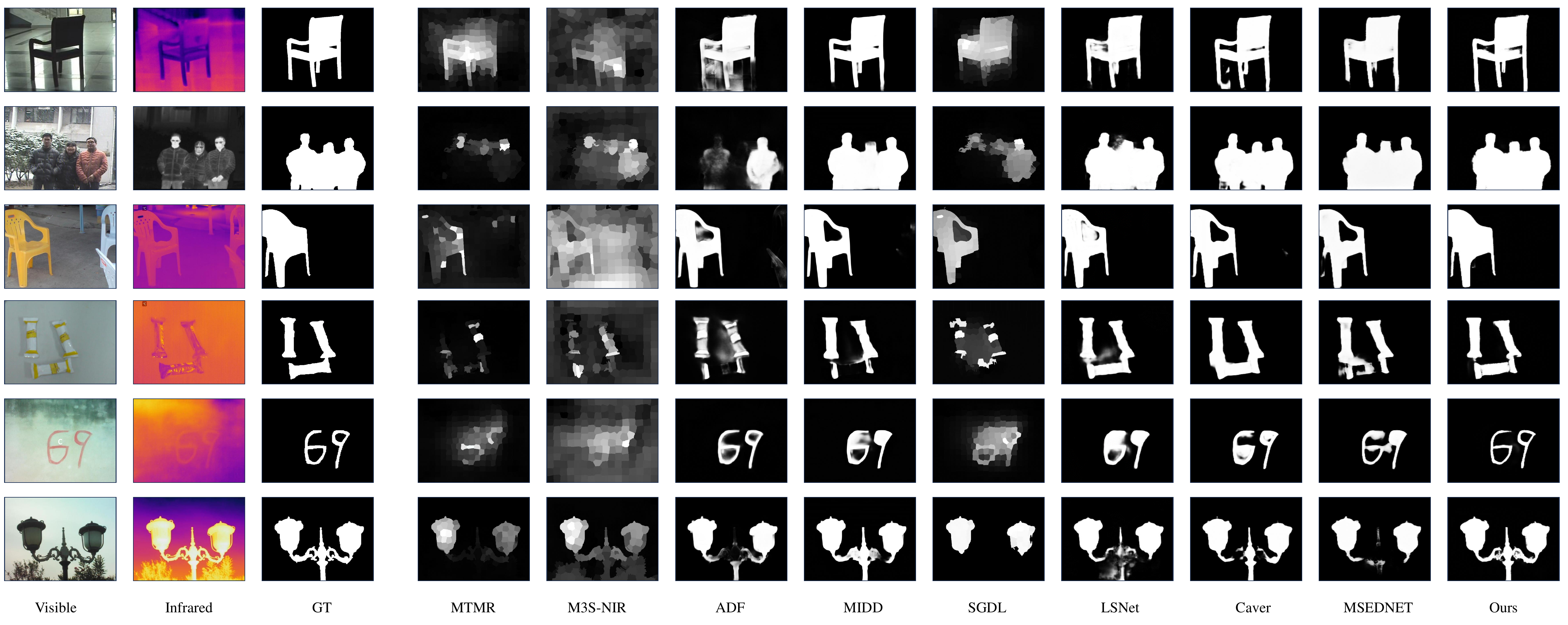}
	\caption{Visual comparison among different SOTA methods and Ours.}
	\label{SOD comparison}
\end{figure*}

\Cref{SOD comparison} presents a comprehensive comparison between our method (last column) and existing approaches for salient object detection. The visualization includes: (1) visible and thermal inputs (first two columns), (2) ground truth masks (third column), and (3) predictions from nine existing methods (remaining columns). Our method demonstrates superior performance across multiple challenging scenarios through three key advantages:
First, the proposed approach effectively leverages cross-modal complementarity between visible and thermal data to produce precise object boundaries with minimal noise interference (rows 1 and 5). 
Second, it exhibits remarkable robustness in occluded scenes (row 2), maintaining complete object morphology where competing methods generate fragmented or blurred detections. Third, our solution shows exceptional sensitivity to small objects (row 4) and complex shapes (rows 3 and 6), achieving detection accuracy that closely matches the ground truth.

Quantitative analysis reveals our method maintains consistent performance across diverse challenging conditions, including low-resolution inputs, occlusions, small objects, and cluttered backgrounds, significantly outperforming existing approaches that show scene-specific limitations. This demonstrates our framework's superior generalization capability and establishes it as a robust solution for salient object detection tasks.

\subsection{Limitations}
\label{subsubsec4.7}
\begin{figure*}[!t]
	\centering
	\includegraphics[width= 0.8\linewidth]{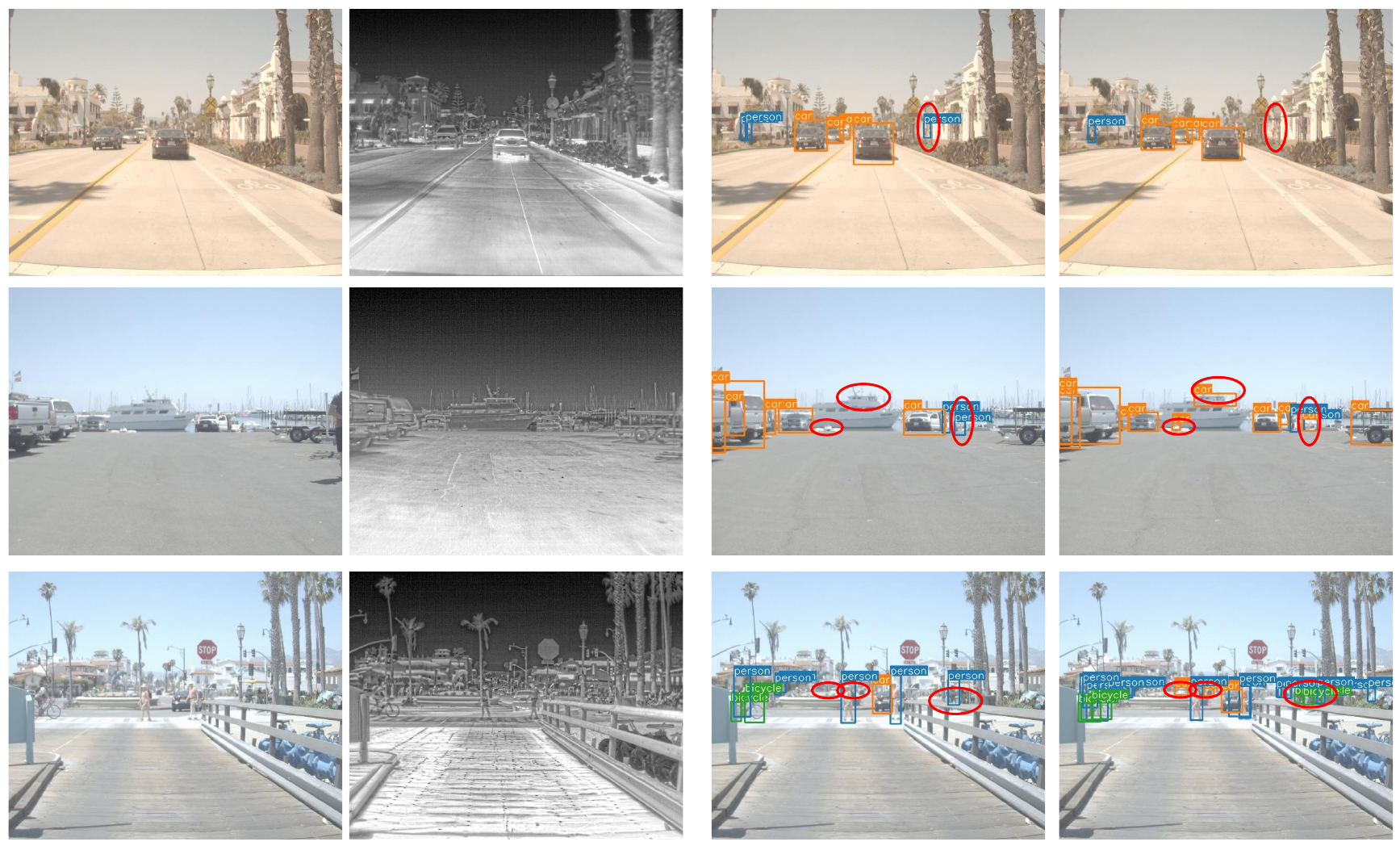}
	\caption{Failure cases on the FLIR dataset. (The first and second columns are the input RGB and thermal image, the third column is groundtruth, and the fourth column is the detection result of our method. The red circles indicate the false positives or false negatives in the images. Zoom in for more details.)}
	\label{false_example}
\end{figure*}
In this section, we analyze several failure cases that highlight the limitations of our method in \Cref{false_example}:
\begin{itemize}
	\item Distant objects: In the first row, the images illustrate scenarios where the object is distant and is not prominent in the thermal image. When the object's thermal signature closely matches the background, our method struggles to accurately distinguish the object from its surroundings.
	\item Occluded objects: The second row of images shows instances where objects are partially or fully occluded. Our method tends to overlook these occluded objects, leading to missed detections. 
	\item False positives with similar thermal appearance: The third row demonstrates situations where objects with thermal characteristics similar to those of a person result in false-positive detections. This challenge is exacerbated by the low resolution of the images, which hampers the model's ability to differentiate between actual objects of interest and irrelevant background features. 
\end{itemize}
These failure cases demonstrate the critical need to improve our model's capability to handle three key challenges: distant small object detection, occluded object detection, and discrimination between thermally similar objects, especially in low-resolution images.

\section{Conclusion}
\label{sec5}
In this paper, we propose MS2Fusion, an adaptive and efficient multispectral feature fusion framework that addresses the key limitations of existing methods. Unlike conventional approaches that rely on rigid fusion strategies or suffer from computational inefficiency, our method leverages dynamic state space models to efficiently capture long-range dependencies while maintaining low computational complexity, overcoming the drawbacks of both CNN-based and Transformer-based techniques. Furthermore, we introduce a shared-parameter state space module to explicitly model cross-modal feature interactions, enhancing discriminative power while mitigating information loss and modal discrepancies. Extensive experiments demonstrate that MS2Fusion achieves state-of-the-art performance, validating its effectiveness in robust multispectral object detection under challenging conditions.

In future work, we plan to adapt MS2Fusion for resource-constrained platforms such as wireless sensor networks and UAV-based systems. Leveraging recent progress in efficient data transmission \citep{dawood2023simulation} and lightweight real-time detection for drones \citep{bakirci2025performance}, we aim to integrate TinyML-oriented optimizations, including model pruning, quantization, and state-space compression, to enhance real-time efficiency and energy awareness. Beyond the current scope, the proposed fusion methodology will be extended to other multimodal sensing tasks under computational and bandwidth constraints, supported by our ongoing construction of a multispectral UAV–maritime dataset for large-scale validation.

\section*{Acknowledgements}

This work was supported in part by the National Natural Science Foundation of China under Grant No. 61903164, the Natural Science Foundation of Jiangsu Province in China under Grants BK20191427 and the Key R\&D Program of Zhejiang Province (2024C04056(CSJ)). SD and HL are not supported by any funds for this work.

\bibliography{ref}


\end{document}